\documentclass[conference]{IEEEtran}
\let\labelindent\relax
\usepackage{times}

\usepackage[numbers]{natbib}
\usepackage{multicol}
\usepackage[bookmarks=true]{hyperref}
\usepackage{array}
\usepackage{hyperref}
\usepackage{amsmath}
\usepackage{cleveref}
\usepackage{paralist}
\usepackage{pdfpages}
\usepackage{enumitem}
\usepackage{wrapfig}
\usepackage{booktabs} 
\usepackage{multirow}
\usepackage{hyperref}
\usepackage{url}
\usepackage{amssymb}
\usepackage{graphicx}
\usepackage{CJKutf8}
\usepackage[noend]{algorithmic}
\usepackage{colortbl}
\usepackage{threeparttable}
\usepackage{mdframed}
\usepackage{tcolorbox}
\usepackage{blindtext}
\usepackage{multicol}
\usepackage{caption, subcaption}
\usepackage[utf8]{inputenc}
\usepackage{tcolorbox}
\usepackage{listings}
\usepackage{xcolor}
\usepackage{tabularx}
\definecolor{customblue}{HTML}{ccf2f5}
\usepackage{soul, color, xcolor}
\soulregister{\cite}7 
\soulregister{\citep}7 
\soulregister{\citet}7 
\soulregister{\ref}7 
\soulregister{\cref}7 
\soulregister{\pageref}7

\newcommand{\ours}[0]{Manual2Skill}

\newcommand{\bluebold}[1]{\cellcolor{customblue}\textbf{#1}}

\begin{document}

\title{Manual2Skill: Learning to Read Manuals and Acquire Robotic Skills for Furniture Assembly Using Vision-Language Models}


\author{
Chenrui Tie$^{* 1}$ \quad
Shengxiang Sun$^{* 2}$ \quad 
Jinxuan Zhu$^{1}$ \quad 
Yiwei Liu$^{4}$ \quad 
Jingxiang Guo$^{1}$ \quad \\
Yue Hu$^{5}$ \quad
Haonan Chen$^{1}$ \quad
Junting Chen$^{1}$ \quad
Ruihai Wu$^{3}$ \quad
Lin Shao$^{1}$ \quad  \\
$^1$National University of Singapore \quad 
\\
$^2$University of Toronto \quad 
$^3$Peking University \quad 
$^4$Sichuan University
$^5$Zhejiang University
}

\vspace{-3mm}
\twocolumn[{%
\renewcommand\twocolumn[1][]{#1}%
\maketitle

\begin{center}
\centering
\captionsetup{type=figure}
\vspace{-10mm}
\includegraphics[page=1, width=\textwidth]{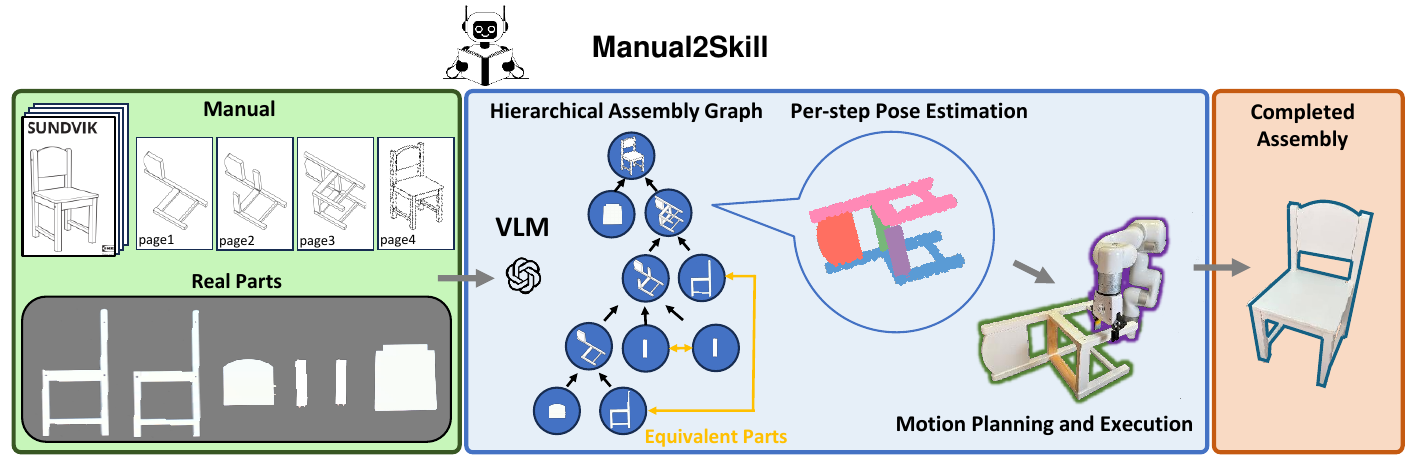} 
\vspace{-5mm}
\caption{\textbf{Overview of Manual2Skill Framework.} We propose Manual2Skill, which learns manipulation skills from manuals, enabling robots to understand and execute complex manipulation tasks in a manner akin to humans. 
The green region showcases the input of our pipeline: the pictures of the assembly manual and real parts.
The blue region depicts our pipeline: 1) a Vision-Language Model (VLM) generates a Hierarchical Assembly Graph, 2) a per-step pose estimation module predicts the 6D-poses of components, and 3) a motion planning and execution module controls the robot arms to assemble the furniture autonomously.}
\label{fig: teaser}
\end{center}
}]
\renewcommand{\thefootnote}
{\fnsymbol{footnote}}
\footnotetext[1]{Equal contribution.}

\begin{abstract}
Humans possess an extraordinary ability to understand and execute complex manipulation tasks by interpreting abstract instruction manuals. For robots, however, this capability remains a substantial challenge, as they cannot interpret abstract instructions and translate them into executable actions.
In this paper, we present \ours, a novel framework that enables robots to perform complex assembly tasks guided by high-level manual instructions. Our approach leverages a Vision-Language Model (VLM) to extract structured information from instructional images and then uses this information to construct hierarchical assembly graphs. These graphs represent parts, subassemblies, and the relationships between them.
To facilitate task execution, a pose estimation model predicts the relative 6D poses of components at each assembly step. At the same time, a motion planning module generates actionable sequences for real-world robotic implementation.
We demonstrate the effectiveness of \ours\ by successfully assembling several real-world IKEA furniture items. This application highlights its ability to manage long-horizon manipulation tasks with both efficiency and precision, significantly enhancing the practicality of robot learning from instruction manuals.
This work marks a step forward in advancing robotic systems capable of understanding and executing complex manipulation tasks in a manner akin to human capabilities.
Project Page: \href{https://owensun2004.github.io/Furniture-Assembly-Web/}{https://owensun2004.github.io/Furniture-Assembly-Web/}
\end{abstract}
\IEEEpeerreviewmaketitle

\section{Introduction}
\label{sec:intro}
Humans can learn manipulation skills from instructions in images or texts; for example, people can assemble IKEA furniture or LEGO models by following a manual's instructions. 
This ability enables humans to efficiently acquire long-horizon manipulation skills from sketched instructions.
In contrast, robots typically learn such skills through imitation learning \cite{10602544} or reinforcement learning \cite{tang2024deep}, both of which require significantly more data and computation. 
Replicating the human ability to transfer abstract manuals to real-world actions remains a significant challenge for robots.
Manuals are typically designed for human understanding, using simple schematic diagrams and symbols to convey manipulation processes. This abstraction makes it difficult for robots to comprehend such instructions and derive actionable manipulation strategies ~\citep{liu2024ikea,wang2022ikea,wang2022translating}.
Developing a method for robots to effectively utilize human-designed manuals would greatly expand their capacity to tackle complex, long-horizon tasks while reducing the demand of collecting extensive demonstration data.

\clearpage


Manuals inherently encode the structural information of complex tasks. They decompose high-level goals into mid-level subgoals and capture task flow and dependencies, such as sequential steps or parallelizable subtasks.
For example, furniture assembly manuals guide the preparation and combination of components and ensure that all steps follow the correct order ~\citep{liu2024ikea}.
Extracting this structure is crucial for robots to replicate human-like understanding and manage complex tasks effectively~\citep{jiang2024roboexp,mo2019structurenet}. 
After decomposing the task, robots need to infer the specific information for each step, such as the involved components and their spatial relationships.
For example, in cooking tasks, the instruction images and texts may involve selecting ingredients, tools, and utensils and arranging them in a specific order ~\citep{shi2023robocook}.
Finally, robots need to generate a sequence of actions to complete the task, such as grasping, placing, and connecting components. 
Previous works have tried to leverage sketched pictures~\citep{sundaresan2024rt} or trajectories~\citep{gu2023rt} to learn manipulation skills but are always limited to relatively simple tabletop tasks.

In this paper, we propose \ours, a novel robot learning framework that is capable of learning manipulation skills from visual instruction manuals.
This framework can be applied to automatically assemble IKEA furniture, a challenging and practical task that requires complex manipulation skills.
As illustrated in~\Cref{fig: teaser}, given a set of manual images and the real furniture parts, we first leverage a vision language model to understand the manual and extract the assembly structure, represented as a hierarchical graph.
Then, we train a model to estimate the assembly poses of all involved components in each step.
Finally, a motion planning module generates action sequences to move selected components to target poses and executes them on robots to assemble the furniture.

In summary, our main contributions are as follows:
\begin{itemize}
    \item We propose \ours, a novel framework that leverages VLM to learn robotic skills from manuals, enabling a generalizable assembly pipeline for IKEA furniture.
    \item We introduce a hierarchical graph generation pipeline that utilizes VLM to extract structured information for assembly tasks. Our pipeline facilitates real-world assembly and extends to other assembly applications.
    \item We define a novel assembly pose estimation task within the learning-from-manual framework. We predict the 6D poses of all involved components at each assembly step to meet real-world assembly requirements.
    \item We evaluate our method on four real items of IKEA furniture, demonstrating its effectiveness and applicability in real-world assembly tasks.
\end{itemize}



\section{Related Work}
\subsection{Furniture Assembly}
Part assembly is a long-standing challenge with extensive research exploring how to construct a complete shape from individual components or parts~\citep{chen2022neural,funkhouser2011learning,jones2021automate,lee2021ikea,li2020learning,scarpellini2024diffassemble,wu2023leveraging,tian2024asap,tian2022assemble}. Broadly, we can categorize part assembly into \emph{geometric assembly} and \emph{semantic assembly}. \emph{Geometric assembly} relies solely on geometric cues, such as surface shapes or edge features, to determine how parts mate together~\citep{chen2022neural,wu2023leveraging,sellan2022breaking,du2024generative}. In contrast, \emph{semantic assembly} primarily leverages high-level semantic information about the parts to guide assembly process~\citep{funkhouser2011learning,jones2021automate,lee2021ikea,li2020learning,tian2022assemble}.

Furniture assembly is a representative \emph{semantic assembly} task, where each part has a predefined semantic role (e.g., a chair leg or a tabletop), and the assembly process follows intuitive, common-sense relationships (e.g., a chair leg must be attached to the chair seat). Previous studies on furniture assembly have tackled different aspects of the problem, including the motion planning~\citep{suarez2018can}, multi-robot collaboration~\citep{knepper2013ikeabot}, and assembly pose estimation~\citep{li2020learning,yu2021roboassembly,li2024category}. Researchers have developed several datasets and simulation environments to facilitate research in this domain. For example,
\citet{wang2022ikea,liu2024ikea} introduced IKEA furniture assembly datasets containing 3D models of furniture and structured assembly procedures derived from instruction manuals. Additionally, \citet{lee2021ikea} and~\citet{yu2021roboassembly} developed simulation environments for IKEA furniture assembly, while
\citet{heo2023furniturebench} provides a reproducible benchmark for real-world furniture assembly.
However, existing works typically focus on specific subproblems rather than addressing the entire assembly pipeline. 
In this work, we aim to develop a comprehensive framework that learns the sequential process of furniture assembly from manuals and deploys it in real-world experiments.

\subsection{VLM Guided Robot Learning}
Vision Language Models (VLMs)~\citep{yin2023survey} have been widely used in robotics to understand the environment~\citep{huang2024copa} and interact with humans~\citep{shi2024yell}.
Recent advancements highlight VLMs' potential to enhance robot learning by integrating vision and language information, enabling robots to perform complex tasks with greater adaptability and efficiency~\cite {huang2024rekep}.
A potential direction is the development of the Vision Language Action Model (VLA Model) that can generate actions based on the vision and language inputs~\citep{black2024pi_0,kim2024openvla,brohan2023rt,team2024octo}.
However, training such models requires vast amounts of data, and they struggle with long-horizon or complex manipulation tasks.
Another direction is to leverage VLMs to guide robot learning by providing high-level instructions and perceptual understanding. VLMs can assist with task descriptions~\citep{huang2024copa,huang2024rekep}, environment comprehension~\citep{jiang2024roboexp}, task planning~\citep{vemprala2024chatgpt,yao2022react,zhao2024large}, and even direct robot control~\citep{li2024manipllm}.
Additionally, \citet{goldberg2024blox} demonstrates how VLMs can assist in designing robot assembly tasks. Building on these insights, we explore how VLMs can interpret abstract manuals and extract structured information to guide robotic skill learning for long-horizon manipulation tasks.

\subsection{Learning from Demonstrations}
Learning from demonstration~(LfD) has achieved promising results in acquiring robot manipulation skills~\citep{fu2024mobile,zhu2023viola,chi2023diffusion}. For a broader review of LfD in robotic assembly, we refer to ~\citet{zhu2018robot}.
The key idea is to learn a policy that imitates the expert's behavior.
However, previous learning methods often require fine-grained demonstrations, like robot trajectories~\citep{chi2023diffusion} or videos~\citep{kareer2024egomimic,sontakke2024roboclip,jonnavittula2024view}.
Collecting these demonstrations is often labor-intensive and may not always be feasible.
Some works propose to learn from coarse-grained demonstrations, like the hand-drawn sketches of desired scenes~\citep{sundaresan2024rt} or rough trajectory sketches~\citep{gu2023rt}.
These approaches reduce dependence on expert demonstrations and improve the practicality of LfD. However, they are mostly limited to tabletop manipulation tasks and do not generalize well to more complex, long-horizon assembly problems.
In this work, we aim to extend LfD beyond these constraints by tackling a more challenging assembly task using abstract instruction manuals. 
\section{Problem Formulation}
\label{sec:method-formulation}
Given a complete set of 3D assembly parts and its assembly manual, our goal is to generate a physically feasible sequence of robotic assembly actions for autonomous furniture assembly. Manuals typically use schematic diagrams and symbols designed to depict step-by-step instructions in an abstract format that is universally understandable. We define the manual pages as a set of $N$ images. $\mathcal{I} = \{I_1, I_2, \cdots, I_N\}$, where each image $I_i$ illustrates a specific step in the assembly process, such as the merging of certain parts or subassemblies.

The furniture consists of $M$ individual parts $\mathcal{P} = \{P_1, P_2, \cdots, P_M\}$. A \emph{part} is an individual element in $\mathcal{P}$ that remains disconnected from other parts until assembly. A \emph{subassembly} is any partially or fully assembled structure that forms a proper subset of $\mathcal{P}$ (for example, $\{P_1, P_2\}$). The term \emph{component} encompasses both parts and subassemblies.

Given the manual and 3D parts, the system generates an assembly plan. Each step corresponds to a manual image and specifies the involved parts and sub-assemblies, their spatial 6D poses, and the assembly actions or motion trajectories required for execution. 
\section{Technical Approach}
Our approach automates furniture assembly by leveraging the VLM to interpret IKEA-style manuals and guide robotic execution. Given a visual manual and physical parts in a pre-assembly scene, a VLM generates a hierarchical assembly graph, defining which parts and subassemblies are involved in each step. Next, a per-step pose estimation model predicts 6D poses for each component using a manual image and the point clouds of involved components. Finally, for assembly execution, the estimated poses are transformed into the robot's world frame, and a motion planner generates a collision-free trajectory for part mating. 

\begin{figure*}[htb!]
\centering
\includegraphics[page=1, width=0.95\textwidth]{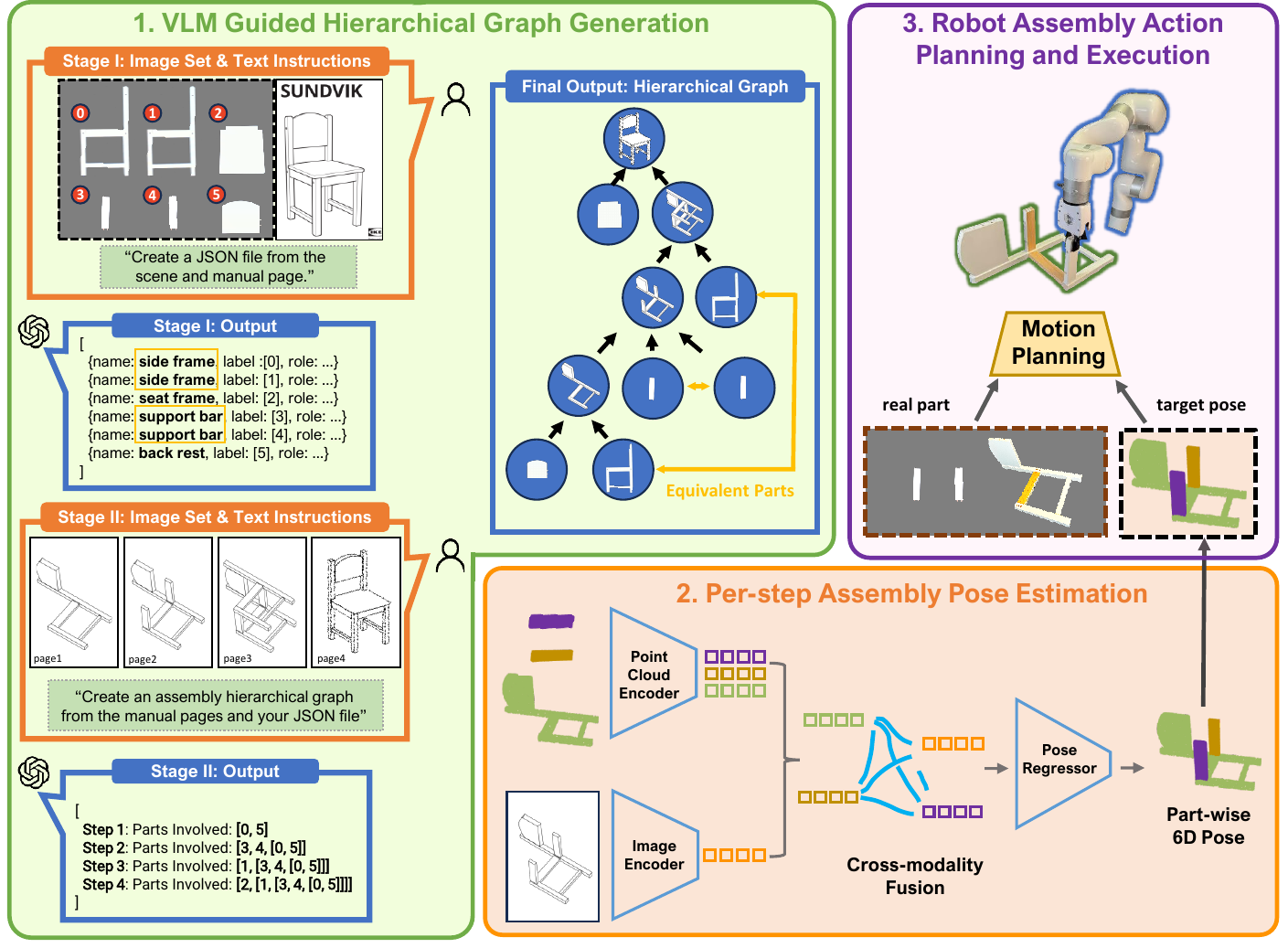} 
\caption{\textbf{Framework Overview.} (1) GPT-4o \cite{achiam2023gpt} is queried with manual pages to generate a sequential assembly plan, represented as a hierarchical assembly graph. 
(2) The furniture components’ point clouds and corresponding manual images are processed by a pose estimation module to predict target poses for each component. (3) The system sequentially executes the assembly by planning and performing robotic actions based on the hierarchical assembly graph and estimated poses.}
\label{fig: pipeline}
\end{figure*}

This paper shows an overview of our framework in Fig.~\ref{fig: pipeline}. We describe the VLM-guided assembly hierarchical graph generation in~\Cref{sec:graph_gen_full_desc}, followed by per-step assembly pose estimation in~\Cref{sec:method-per-step} and assembly action generation based on component relationships in~\Cref{sec:method-action}.

\subsection{VLM Guided Hierarchical Assembly Graph Generation}
\label{sec:graph_gen_full_desc}

This section demonstrates how VLMs can interpret IKEA-styled manuals to generate high-level assembly plans. Given a manual and a real-world image of furniture parts (\emph{pre-assembly scene image}), a VLM predicts \emph{a hierarchical assembly graph}. We show one example in Fig.~\ref{fig: pipeline}. In this graph, leaf nodes represent atomic parts, while non-leaf nodes denote subassemblies. We structure the graph in multiple layers, where each layer contains nodes representing parts or subassemblies involved in a single assembly step (corresponding to one manual image). The directed edges from the children to a parent node indicate that the system assembles the parent node from all its children nodes. 
Additionally, we add edges between equivalent parts, denoting these parts are identical($e.g.$ four legs of a chair).
Representing the assembly process as a hierarchical graph can decomposes the assembly into sequential steps while specifying necessary parts and subassemblies. 
We give the formal definition of the hierarchical graph in~\Cref{appd:def_assembly_graph}.
We achieve this in two stages: \textit{Associating Manuals with Real Parts} and \textit{Identifying Parts needed in Each Image}.

\subsubsection{VLM Capabilities and General Prompt Structure} The task is inherently complex due to the diverse nature of input images. Manuals are typically abstract sketches, whereas \emph{pre-assembly scene images} are high-resolution real-world images. 
Such diversity requires advanced visual recognition and spatial reasoning across varied image domains, which are strengths of VLMs due to their training on extensive, internet-scale datasets. We demonstrate the effectiveness of VLMs for this task in \Cref{assembly_tree_gen} and 
\Cref{vlm_plan_complete}.

Every VLM prompt consists of two components: 
\begin{itemize}
  \item  \textbf{Image Set:} This includes all manual pages and the real-world \emph{pre-assembly scene image}. Unlike traditional VLM applications in robotics \cite{kim2024openvla, huang2024rekep}, which process a single image, our method requires multi-image reasoning.
  \item \textbf{Text Instructions:} These instructions provide a task-specific context, guiding the model in interpreting the image set. The instructions range from simple directives to Chain-of-Thought reasoning \cite{wei2022chain}. All instructions incorporate in-context learning examples, specifying the required output format—be it JSON, Python code, or natural language. This structure is essential to our multi-stage pipeline, ensuring well-structured, interpretable outputs that seamlessly integrate into subsequent stages.
\end{itemize}

\subsubsection{Stage I: Associating Real Parts with Manuals} 
Given the manual's cover sketch of the assembled furniture and the \emph{pre-assembly scene image}, the VLM aims to associate physical parts with the manual. The VLM achieves this by predicting the roles of each physical part through semantically interpreting the manual's illustrations. This process involves analyzing spatial, contextual, and functional cues in the manual illustrations to enable a comprehensive understanding of each physical part.
Our method follows CoT \cite{wei2022chain} and Least-to-Most \cite{zhou2022least} prompting 
for better accuracy. 

To enhance part identification, we employ Set of Marks \cite{yang2023set} and GroundingDINO \cite{liu2025grounding} to automatically label parts on the \emph{pre-assembly scene image} with numerical indices. The labeled scene image and manual sketch form the \textbf{Image Set}. \textbf{Text instructions} consist of a brief context explanation for the association task of predicting the roles of each physical part, accompanied by in-context examples of the output structure: \\
\centerline{\emph{\{name, label, role\}}}

For example, in~\Cref{fig: pipeline} In Stage I Output, we describe the chair's seat as \emph{{name: seat frame, label: [2], role: for people sitting on a chair, the seat offers essential support and comfort and is positioned centrally within the chair's frame}}. Here, `\emph{[2]}' indicates that this triplet corresponds to the physical part labeled with index 2 in the pre-assembly scene image. This triplet format enhances interpretability and ensures consistency by structuring all outputs into the same data format. We use the Image Set and Text Instructions as the input prompt for the VLM (specifically GPT-4o \cite{achiam2023gpt}) and query it once to generate real assignments for all physical parts. We then use these labels as leaf nodes in the hierarchical assembly graph.

We can obtain equivalent parts through these triplets. When two physical parts share the same geometric shapes, their triplets only differ by label. For example, in~\Cref{fig: pipeline} Stage I Output, \emph{\{name: side frame, label: [0], role:...\}} and \emph{\{name: side frame, label: [1], role:...\}}—these two parts are considered equivalent. Understanding equivalent part relationships is crucial for downstream modules, as demonstrated by our ablation experiments(see~\Cref{appd:pose_ablation}).

\subsubsection{Stage II: Identify Involved Parts in Each Step}
This stage focuses on identifying the particular parts and subassemblies involved in each manual page. The VLM achieves this by reasoning through the illustrated assembly steps, using the triplets and the labeled pre-assembly scene from the previous stage as supporting hints.

In practice, we observe that irrelevant elements in the manual (e.g., nails, human figures) can distract the VLM. 
We use the cropped manual images from~\cite{wang2022ikea}, where only the furniture parts and subassemblies are reserved to focus the VLM's attention (\Cref{fig: pipeline} Stage II Image Set), significantly improving performance (see~\Cref{appd:VLM_ablation} for details). 

The manual pages, combined with the labeled pre-assembly scene from the previous stage, form the \textbf{Image Set}. 
The \textbf{Text Instructions} use a Chain-of-Thought prompt to guide the VLM in identifying parts and subassemblies step by step and includes in-context examples that clarify the structured output format: a pair consisting of (Step N, Labeled Parts Involved). 
The bottom left output of~\Cref{fig: pipeline} provides an example of this format.
Together, the \textbf{Image Set} and \textbf{Text Instructions} compose the input prompt for GPT-4o, which generates pairs for all assembly steps using a single query. 

As shown in Fig.~\ref{fig: pipeline}, the system outputs nested lists. We then transform these lists, along with the equivalent parts, into a hierarchical graph. Using this assembly graph, we traverse all non-leaf nodes and explore various assembly orders. Formally, a feasible assembly order is an ordered set of non-leaf nodes, ensuring that a parent node appears only after all its child nodes. A key advantage of the hierarchical graph representation is its flexibility—since the assembly sequence is not unique, it allows for parallel assembly or strategic sequencing.

\subsection{Per-step Assembly Pose Estimation}
\label{sec:method-per-step}

Given an assembly order, we train a model to estimate the poses of components (parts or subassemblies) at each step of the assembly process. At each step, the model inputs the manual image and the point clouds of the involved components, predicting their target poses to ensure proper alignment. To support this task, we construct a dataset for sequential pose estimation. For a detailed description, see~\Cref{appd: Per-step Assembly Pose Estimation Dataset}.

Given each component's point cloud (obtained from real-world scans or our dataset), we first center it by translating its centroid to the origin. Next, we apply Principal Component Analysis (PCA) to identify the dominant object axes, which define a canonical coordinate frame. The most dominant axes serve as the reference frame, ensuring a shape-driven and consistent orientation that remains independent of arbitrary coordinate systems.

The dataset we create provides manual images, point clouds, and target poses for each component in the camera frame of the corresponding manual image(following~\cite{li2020learning}). For an assembly step depicted in the manual image $\mathcal{I}_i$, the inputs to our model include: (1) the manual image $\mathcal{I}_i$; (2) the point clouds of all involved components. The output is the target pose $T\in SE(3)$ for each component represented in the camera frame of $I_i$. 

\subsubsection{Model Architecture} Note that the number of components at each step is not fixed, depending on the subassembly division of the furniture.
Our pose estimation model consists of four parts: an image encoder $\mathcal{E}_I$, a point cloud encoder $\mathcal{E}_P$, a cross-modality fusion module $\mathcal{E}_G$, and a pose regressor $\mathcal{R}$.

We first feed the manual image $I$ into the image encoder to get an image feature map $\mathbf{F}_I$.
\begin{equation}
    \mathbf{F}_I = \mathcal{E}_I(I)
\end{equation}
Then, we feed the point clouds into the point cloud encoder to get the point cloud feature for each component. 
\begin{equation}
    \{\mathbf{F}_j\} = \mathcal{E}_P(\{P\}_j)
\end{equation}

In order to fuse the multi-modality information from the manual image and the point cloud features, we leverage a GNN~\citep{wu2020comprehensive} to update the information for each component.
We consider the manual image feature and component-wise point cloud features as nodes in a complete graph, employing a GNN to update the information for each node.
\begin{equation}
    \mathbf{F}_I',\{\mathbf{F}_j'\} = \mathcal{E}_G(\mathbf{F}_I,\{\mathbf{F}_j\})
\end{equation}
where $\mathbf{F}_I',\{\mathbf{F}_j'\}$ are updated image and point cloud features.

Finally, we feed the updated point cloud features as input into the pose regressor to get the target pose for each component.
\begin{equation}
    T_j = \mathcal{R}(\mathbf{F}_j')
\end{equation}

\subsubsection{Loss Function} 
We adopt a loss function that jointly considers pose prediction accuracy and point cloud alignment, following~\citep{zhang2024manual,li2024category}. The first term penalizes errors in the predicted SE(3) transformation, while the second measures the distance between predicted and ground truth point clouds. To account for interchangeable components, we compute the loss across all possible permutations of equivalent parts and select the minimum loss as the final training objective. We provide further details on the loss formulation and training strategy in~\Cref{appd:implementation_pose}.

\subsection{Robot Assembly Action Generation}  
\label{sec:method-action}  

\subsubsection{Align Predicted Poses with the World Frame}
At each assembly step, the previous stage predicts each component's pose in the camera frame of the manual image. However, real-world robotic systems operate in their world frame, requiring a 6D transformation between these coordinates. Consider two components, A and B. The predicted target poses in the camera frame are denoted as $^{I_i}\hat{\mathcal{T}}_a$ and $^{I_i}\hat{\mathcal{T}}_b$. Meanwhile, our system can collect the current 6D pose of part A in the world frame, represented as $^{W}\mathcal{T}_a$. 
To align $^{I_i}\hat{\mathcal{T}}_a$ to $^{W}\mathcal{T}_a$, we compute the 6D transformation matrix ${}^{W}_{I_i}\mathcal{T}$, which maps the camera frame to the world frame.
\begin{equation}
    ^{W}\mathcal{T}_a = {}^{W}_{I_i}\mathcal{T}  {}^{I_i}\hat{\mathcal{T}}_a
\end{equation}
Using the same transformation ${}^{W}_{I_i}\mathcal{T}$, we compute the assembled target pose of part B (and all remaining components) in the world frame.
\begin{equation}
    ^{W}\mathcal{T}_b = {}^{W}_{I_i}\mathcal{T}  {}^{I_i}\hat{\mathcal{T}}_b
\end{equation}
This transformation accurately maps predicted poses from the manual image frame to the robot's world frame, ensuring precise assembly execution.

\subsubsection{Assembly Execution}
Once our system determines the target poses of each component in the world frame for the current assembly step, it grasps each component and generates the required action sequences for assembly.

\paragraph{Part Grasping}
After scanning each real-world part, we obtain the corresponding 3D meshes for each part. We employ FoundationPose~\cite{wen2024foundationpose}, and the Segment Anything Model (SAM)~\cite{kirillov2023segany} to obtain the initial poses of all parts in the scene.

Given the pose and shape of each part, we design heuristic grasping methods tailored to the geometry of individual components. While general grasping algorithms such as GraspNet~\cite{fang2023anygrasp} are viable, grasping is beyond the scope of this work. Instead, we employ heuristic grasping strategies specifically designed for structured components in assembly tasks. For stick-shaped components, we grasp the centroid of the object after identifying its longest axis for stability. For flat and thin-shaped components, we use fixtures or staging platforms to securely position the object, allowing the robot to grasp along the thin boundary for improved stability. We provide further details on these grasping methods in~\Cref{appd: real-world}.

\paragraph{Part Assembly Trajectory}  
\label{sec:method-realworld-assemble}  

Once the robot arm grasps a component, it finds a feasible, collision-free path to predefined robot poses (anchor poses). At these poses, the 6D pose of the grasped component is recalculated in the world frame, leveraging the FoundationPose ~\cite{wen2024foundationpose} and the Segment Anything Model (SAM)\cite{kirillov2023segany}. The system then plans a collision-free trajectory to the component's target pose. We use RRT-Connect~\cite{kuffner2000rrt} as our motion planning algorithm. All collision objects in the scene are represented as point clouds and fed into the planner. Once the planner finds a collision-free path, the robot moves along the planned trajectory.

\paragraph{Assembly Insertion Policy}
Once the robot arm moves a component near its target pose, the assembly insertion process begins. Assembly insertions are contact-rich tasks that require multi-modal sensing (e.g., force sensors and closed-loop control) to ensure precise alignment and secure connections. However, developing closed-loop assembly insertion skills is beyond the scope of this work and will be addressed in future research. In our current approach, human experts manually perform the insertion action.

\section{Experiments}
In this section, we perform a series of experiments aimed
at addressing the following questions.
\begin{itemize}
    \item Q1: Can our proposed hierarchical assembly graph generation module effectively extract structured information from manuals? (see~\Cref{assembly_tree_gen}) 
     \item Q2: Can the per-step pose estimation be applicable to different categories of furniture and outperform previous settings? (see~\Cref{pose_exp}) 
     \item Q3: How effective is the proposed framework in the assembly of furniture with manual guidance? (see~\Cref{sec:Q1_exp}) 
     \item Q4: Can this pipeline be applied to real-world scenarios?(see~\Cref{real_world_exp})
     \item Q5: Can this pipeline be extended to other assembly tasks? (see~\Cref{gen_other_task})
     \item Q6: How should we determine and evaluate the key design choices of each module? (ablation experiments, see~\Cref{appd:VLM_ablation,appd:pose_ablation})
\end{itemize}
In addition, we have included a comprehensive set of prompts utilized in the VLM-guided hierarchical graph generation process in~\Cref{appd: additional_prompts}

\subsection{Hierarchical Assembly Graph Generation} \label{assembly_tree_gen}
In this section, we evaluate the performance of our VLM-guided hierarchical assembly graph generation approach. Specifically, we assess Stage II: Identifying Parts in Each Image using the IKEA-Manuals dataset~\cite{wang2022ikea}. We provide the rationale for excluding Stage I evaluation in~\Cref{appd: no_stage1}.


\begin{table}[htb!]
    \centering
    \setlength\tabcolsep{4pt} 
    \renewcommand{\arraystretch}{1.2} 
    \begin{threeparttable}
    \captionsetup{width=\linewidth}
    \caption{\textbf{Success Rate Across Task Complexity} ($\uparrow$)}
    \label{tab:parts_performance}
    \begin{tabular}{@{}lcccccc@{}} 
    \toprule
    Number of Parts & 2$\sim$4 & 5$\sim$6 & 7$\sim$8 & 9+ & Average \\ 
    \midrule
    SingleStep & 35.7 & 16.7 & 0 & 0 & 10.8 \\
    GeoCluster & 21.4 & 2.8 & 0 & 0 & 3.9 \\
    Ours (Scene Variations) & \textbf{78.6} & 50.0 & \textbf{12.5} & 0 & 31.4\\ 
    \rowcolor{customblue} \textbf{Ours} & \textbf{78.6} & \textbf{55.6} & 8.3 & 0 & \textbf{32.4}\\ 
    \midrule
    Furniture Count & 14 & 36 & 24 & 28  \\ 
    \bottomrule
    \label{tab:assembly_plan}
    \end{tabular}
    \end{threeparttable}
    \vspace{-2mm}
\end{table}

\begin{figure}[!htb]
    \centering
    \includegraphics[width=1.0\linewidth]{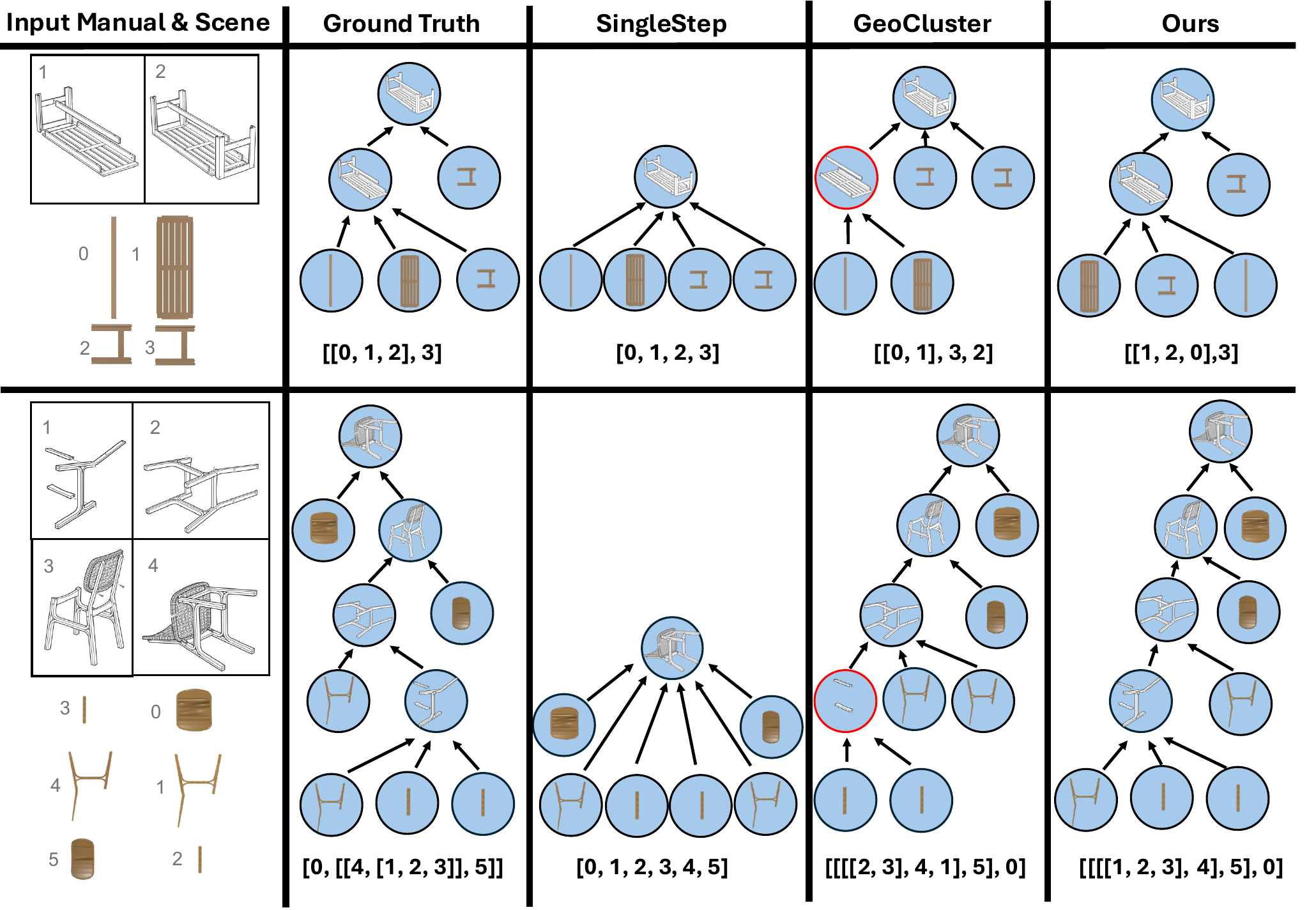}
    \caption{\textbf{Qualitative results.} Our method significantly outperforms the baselines. SingleStep fails on moderately complex furniture, while GeoCluster generates physically impossible subassemblies (highlighted in red). In contrast, our approach closely aligns with the ground truth.}
    \label{fig:tree_viz}
\end{figure}

\begin{figure}[!htb]
    \centering
    \includegraphics[width=1.0\linewidth]{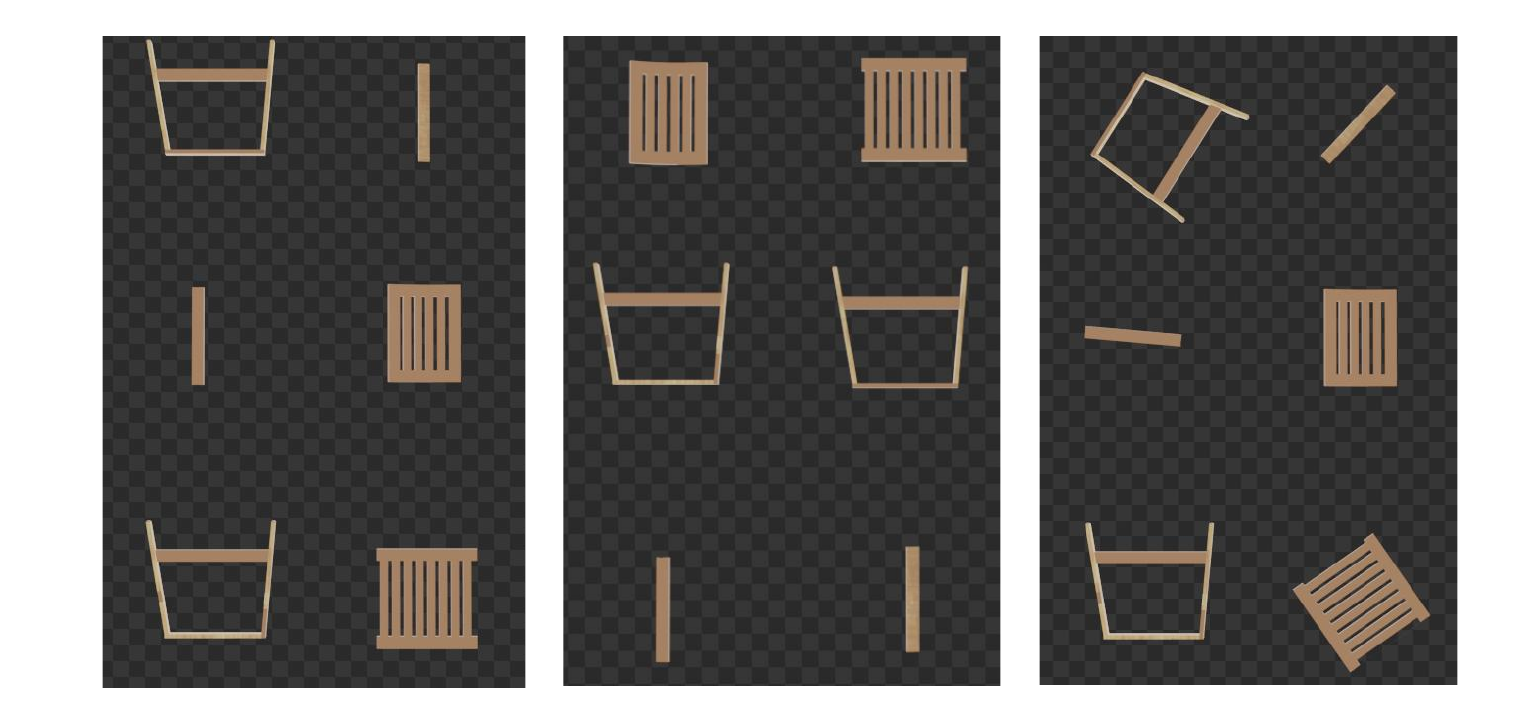}
    \caption{\textbf{Pre-Assembly Scene Variations.} (Left) original pre-assembly scene. (Middle) parts randomly shuffled along the ground plane. (Right) parts randomly rotated in-place.}
    \label{fig:pre_var}
\end{figure}

\textbf{Experiment Setup.} The IKEA-Manuals dataset~\cite{wang2022ikea} includes 102 furniture items, each with IKEA manuals, 3D parts, and assembly plans represented as trees in nested lists. For each item, we load its 3D parts into Blender and render two images: one depicting the original pre-assembly scene with neatly arranged parts, and another showing a scene variation where parts are arbitrarily perturbed (e.g., rotated and shuffled) along a ground plane (see \Cref{fig:pre_var}). This randomization introduces diversity, better simulating real-world pre-assembly scenarios where parts may be disorganized.
 
Each rendered image of the pre-assembly scene, along with the manual, is processed by the VLM through the stages outlined in~\Cref{sec:graph_gen_full_desc} to generate a hierarchical assembly graph. Since we represent our graph as a nested list, we align our notation with the assembly tree notation used in IKEA-Manuals \cite{wang2022ikea}. In this subsection, we refer to our generated assembly graph as the \textit{predicted tree}.

\textbf{Evaluation Metric.} 
We use \textit{Success Rate} criterion, which measures the proportion of the predicted tree that exactly matches the ground-truth tree. We consider a predicted tree exactly matched if all its nodes contain the same set of children nodes as their ground-truth counterparts.
All equivalent parts are seen as identical when computing the tree matching.




\textbf{Baselines.} We compare our VLM-based method against two heuristic approaches introduced in IKEA-Manuals~\cite{wang2022ikea}.

\begin{itemize}
    \item SingleStep predicts a flat, one-level tree with a single parent node and $n$ leaf nodes.
    \item GeoCluster employs a pre-trained DGCNN \cite{wang2019dynamic} to iteratively group furniture parts with similar geometric features into a single assembly step. Compared to SingleStep, it generates deeper trees with more parent nodes and multiple hierarchical levels.
\end{itemize}


\textbf{Results.} As shown in \Cref{tab:assembly_plan}, understanding and interpreting manuals is a challenging task. Our VLM‑guided approach effectively handles furniture manuals with up to 6 parts, a significant breakthrough where baselines struggle even in simpler cases. This $\le6$ parts threshold reflects current VLM capabilities in complex visual-spatial reasoning tasks. Our framework is designed to be scalable, benefiting from rapid VLM advancements. 
We anticipate performance improvements as more powerful VLMs emerge. \Cref{tab:assembly_plan} also highlights the VLM's generalization ability: pre-assembly scene variations in \Cref{fig:pre_var} minimally impact assembly graph generation performance, with average success rates drop within 1\% across all 102 furniture items. \Cref{fig:tree_viz} provides qualitative results for two furniture items, illustrating the advantages of our approach in greater detail.

For ablation studies on the necessity of using and segmenting manuals, see \Cref{appd:VLM_ablation}. Failure cases are further explained in \Cref{appd: failure cases}. Additional results and prompt templates are in \Cref{vlm_plan_complete,appd: additional_prompts}, respectively.

\subsection{Per-step Assembly Pose Estimation}\label{pose_exp}

\begin{table*}[ht]
\centering
\begin{threeparttable}
\caption{\textbf{Quantitative Results of Pose Estimation.}}
\label{tab:pose_estimation}
\renewcommand{\arraystretch}{1.2} 
\begin{tabular}{@{}l>{\centering\arraybackslash}p{0.8cm} >{\centering\arraybackslash}p{0.8cm} >{\centering\arraybackslash}p{0.8cm} >{\centering\arraybackslash}p{0.8cm} >{\centering\arraybackslash}p{0.8cm} >{\centering\arraybackslash}p{0.8cm} >{\centering\arraybackslash}p{0.8cm} >{\centering\arraybackslash}p{0.8cm} >{\centering\arraybackslash}p{0.8cm} >{\centering\arraybackslash}p{0.8cm} >{\centering\arraybackslash}p{0.8cm} >{\centering\arraybackslash}p{0.8cm} >{\centering\arraybackslash}p{0.8cm} >{\centering\arraybackslash}p{0.8cm} >{\centering\arraybackslash}p{0.8cm}@{}}
\toprule

& \multicolumn{3}{c}{GD$\downarrow$} & \multicolumn{3}{c}{RMSE$\downarrow$} & \multicolumn{3}{c}{CD $\downarrow$} & \multicolumn{3}{c}{PA$\uparrow$} \\

\cmidrule(lr){2-4}  \cmidrule(lr){5-7} \cmidrule(lr){8-10} \cmidrule(lr){11-13}

Method & Chair & Lamp & Table & Chair & Lamp & Table & Chair & Lamp & Table & Chair & Lamp & Table\\

\midrule
 \citet{li2020learning} &1.847 &1.865 &1.894 &0.247 &0.278 &0.318 &0.243 &0.396 &0.519 &0.268 &0.121 &0.055\\
Mean-Max Pool &0.434 &1.118 &1.059 &0.087 &0.187 &0.200 &0.046 &0.229 &0.280 &0.457 &0.199 &0.107\\
\textbf{Ours}   & 
\cellcolor{customblue} \textbf{0.202} & 
\cellcolor{customblue} \textbf{0.826} & 
\cellcolor{customblue} \textbf{0.953} & 

\cellcolor{customblue} \textbf{0.042} & 
\cellcolor{customblue} \textbf{0.153} & 
\cellcolor{customblue} \textbf{0.172} & 

\cellcolor{customblue} \textbf{0.027} &  
\cellcolor{customblue} \textbf{0.189} & 
\cellcolor{customblue} \textbf{0.276} & 

\cellcolor{customblue} \textbf{0.868} &  
\cellcolor{customblue} \textbf{0.240} & 
\cellcolor{customblue} \textbf{0.184} \\

\bottomrule
\end{tabular}
\end{threeparttable}
\end{table*}

\textbf{Data Preparation.}
We select three categories of furniture items from PartNet~\citep{mo2019partnet}: chair, table, and lamp.
For each category, we select 100 furniture items and generate 10 parts selection and subassembly division for each piece of furniture.
To generate the assembly manual images, we render diagrammatic images of parts at 20 random camera poses using Blender's Freestyle functionality.
We provide more details about it in~\Cref{appd: Per-step Assembly Pose Estimation Dataset}.
In general, we generate 12,000 training and 5,200 testing data pieces for each category.

\textbf{Training Details.}
For the Image Encoder $\mathcal{E}_I$, we selected the encoder component of DeepLabV3+, which includes MobileNet V2 as the backbone and the atrous spatial pyramid pooling (ASPP) module. We made this choice because DeepLabV3+ leverages atrous convolutions on the basis of Auto Encoder, enabling the model to capture multi-scale structures and spatial information effectively~\citep{chen2017deeplab,chen2018encoder}. It generates a multi-channel feature map from the image $I$, and we use mean-max pool~\citep{zhang2018learning} to derive a global vector $\mathbf{F}_I \in \mathbb{R}^{256}$ from the feature map.
For the Point Clouds Encoder $\mathcal{E}_P$, we use the encoder part of PointNet++~\citep{qi2017pointnet++}. 
For each part and subassembly, we extract a part-wise feature $\mathbf{F}_j \in \mathbb{R}^{256}$.
For the GNN $\mathcal{E}_G$, we use a three-layer graph transformer~\citep{Costa2021.06.02.446809}.
The pose regressor $\mathcal{R}$ is a three-layer MLP. We provide more details of the mean-max pool for the image feature and our training hyperparameter setting in~\Cref{appd:implementation_pose}.

\textbf{Baselines.}
We evaluate the performance of our method on our proposed per-step assembly pose estimation dataset. 
We compare our method with two baselines: 
\begin{itemize}
    \item \citet{li2020learning} proposed a pipeline for single image guided 3D object pose estimation. 
    \item Mean-Max Pool is a variant of our method, replacing GNN with a mean-max pool trick, similar to our approach of obtaining a one-dimensional vector from a multi-channel feature map, with details in~\Cref{appd:implementation_pose}.

\end{itemize}

\textbf{Evaluation Metrics.}
We adopt comprehensive evaluation metrics to assess the performance of our method and baselines.
\begin{itemize}
    \item Geodesic Distance (GD), which measures the shortest path distance on the unit sphere between the predicted and ground-truth rotations.
    \item Root Mean Squared Error (RMSE), which measures the Euclidean distance between the predicted and ground-truth poses.
    \item Chamfer Distance (CD), which calculates the holistic distance between the predicted and the ground-truth point clouds.
    \item Part Accuracy (PA), which computes the Chamfer Distance between the predicted and the ground truth point clouds; if the distance is smaller than 0.01m, we count this part as ``correctly placed".
\end{itemize}

\begin{figure}[!ht]
    \centering
    \begin{minipage}{\linewidth}
    \includegraphics[width=\textwidth]{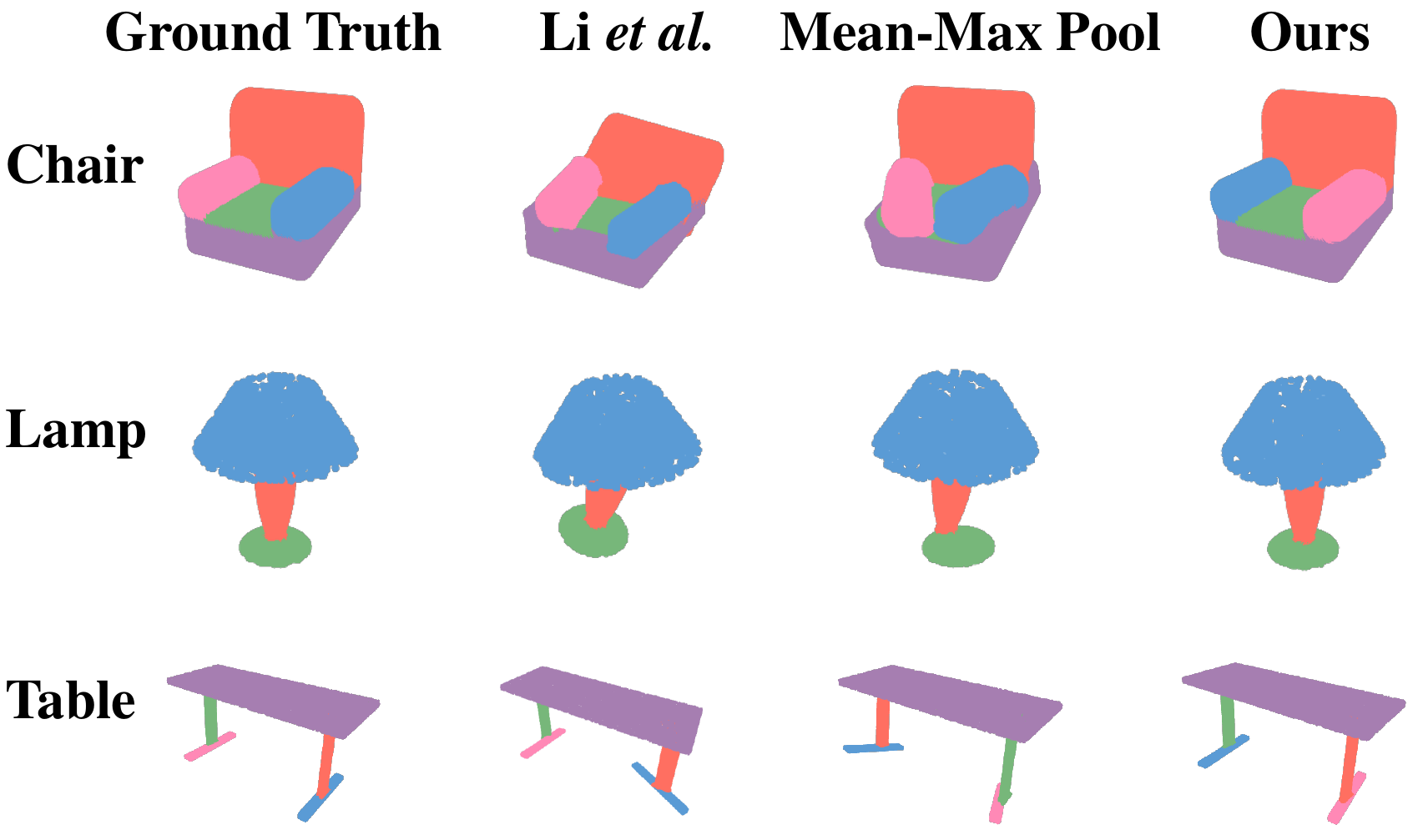}
    \caption{\textbf{Qualitative results on three furniture categories.} We observe better pose predictions than baselines.}
    \label{fig:pose_estimation_viz}
    \end{minipage}
\end{figure}

\textbf{Results.}
As shown in~\Cref{tab:pose_estimation}, our method outperforms ~\citet{li2020learning} and the mean-max pool variant in all evaluation metrics and on three furniture categories.
We attribute this to the effectiveness of our multi-modal feature fusion and GNN in capturing the spatial relationships between parts.
We also provide qualitative results for each furniture category in~\Cref{fig:pose_estimation_viz}.

\textbf{Ablation.}
To assess the impact of equivalent parts, guided image, and per-step data about subassemblies, we perform ablation studies on these components. We present the details and results in~\Cref{appd:pose_ablation}.

\begin{figure*}[htb!]
    \centering
    \begin{minipage}{\linewidth}
        \centering
        \includegraphics[width=0.95\textwidth]{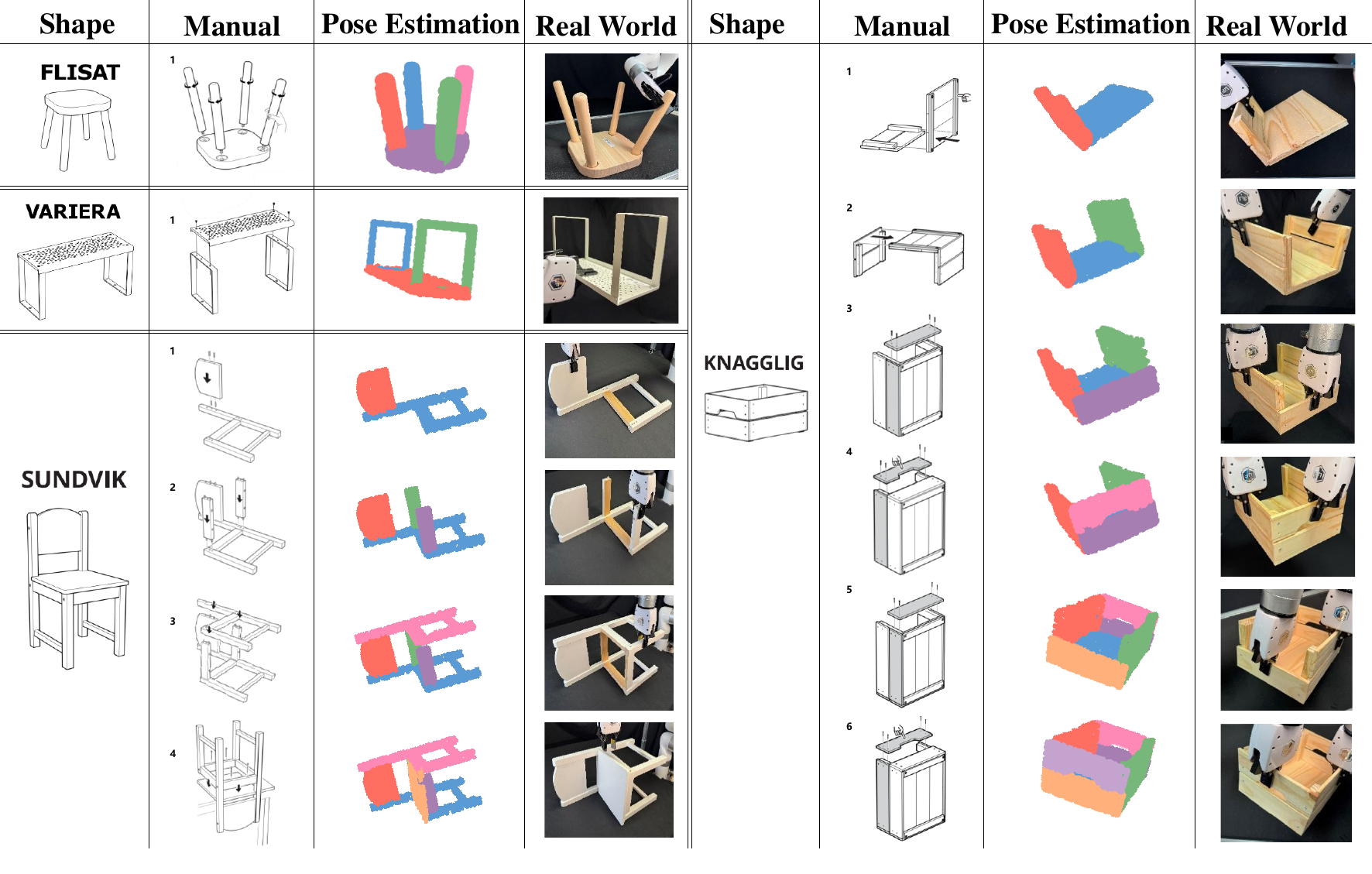}
        \caption{\textbf{Qualitative Evaluation on real IKEA furniture items.} This figure illustrates the assembly process of various IKEA furniture items, including FLISAT, VARIERA, SUNDVIK, and KNAGGLIG, with our approach. For each item, we display the manual images, per-step 3D parts pose estimation results, and real-world assembly outcomes.}
        \label{fig:real_assembly_process}
    \end{minipage}
\end{figure*}

\subsection{Overall Performance Evaluation}\label{sec:Q1_exp}

We evaluate the overall performance of our method by assembling furniture models in a simulation environment. We implement the evaluation process in the PyBullet~\citep{coumans2015bullet} simulation environment and test the entire pipeline. We source all test furniture models from the IKEA-Manuals dataset~\citep{wang2022ikea}. Given these manuals along with 3D parts, we generate the pre-assembly scene images as described in ~\ref{sec:method-action}, and our pipeline generates the hierarchical graphs. Then, we traverse the hierarchical graph to determine the assembly order. Following this sequence and the predicted 6D poses of each component, we implement RRT-Connect~\cite{kuffner2000rrt} in simulation to plan feasible motion paths for the 3D parts and subassemblies, ensuring they move towards their target poses. Note that, in this experiment, we focus on object-centric motion planning and omit robotic execution in our framework. 

\textbf{Baselines.}
As the first to propose a comprehensive pipeline for furniture assembly, there is no direct baseline for comparison.
So we design a baseline method that uses previous work ~\citep{li2020learning} to estimate the poses of all parts, with the guidance of an image of the fully assembled furniture, and adopt a heuristic order to assemble all parts. Specifically, given the predicted poses of all parts, we can calculate the distance between each pair of parts. The heuristic order is defined as follows: starting from a random part, we find the nearest part to it and assemble it, then successively find the nearest part to the assembled parts until we assemble all parts.

\textbf{Evaluation Metrics.}
We adopt the assembly success rate as the evaluation metric and define the following situations as a failure:
1) A part is placed at a pose that is too far from the ground truth pose.
2) A part collides with other parts when moving to the estimated pose. In other words, the RRT-Connect algorithm~\cite{kuffner2000rrt} finds no feasible path when mating it with other parts.
3) We place a part that is not near any other components, causing it to suspend in midair after each assembly step.

\begin{table}[htb!]
    \centering
    \setlength\tabcolsep{4pt}
    \begin{threeparttable}
    \captionsetup{width=\linewidth}
    \caption{\textbf{Success Rate on 4 Furniture Categories$(\uparrow)$}}
    \label{tab: overall}
    \vspace{2mm} 
    \begin{tabular}{lccccc}
    \toprule
     Method & Bench & Chair & Table & Misc & Average \\ 
    \midrule
    \citet{li2020learning}+Heuristic & 0.00&0.39&0.11&0.00&0.30\\
    \textbf{Ours} & 
    \cellcolor{customblue}\textbf{0.67}&
    \cellcolor{customblue}\textbf{0.61}&
    \cellcolor{customblue}\textbf{0.44}&
    \cellcolor{customblue}\textbf{0.50}&
    \cellcolor{customblue}\textbf{0.58}\\
    \bottomrule
    \end{tabular}
    \vspace{-2mm} 
    \end{threeparttable}
\end{table}

\textbf{Results.} We evaluate the overall performance on 50 furniture items from the IKEA-Manual dataset~\citep{wang2022ikea}, each consisting of fewer than seven parts. These items fall into four categories (Bench, Chair, Table, Misc), and we report the success rate for each in~\Cref{tab: overall}.

Our system successfully assembles 29 out of 50 furniture pieces, whereas the baseline method assembles only 15. Our framework achieves a success rate of \textbf{58\%}, demonstrating the effectiveness of our proposed framework. The most common failure occurs when the VLM fails to generate a fully accurate assembly graph, leading to misalignment between the point cloud and the instruction manual images used for pose estimation.

\begin{figure}[thb!]
    \centering
    \resizebox{0.45\textwidth}{0.2\textheight}{
    \includegraphics{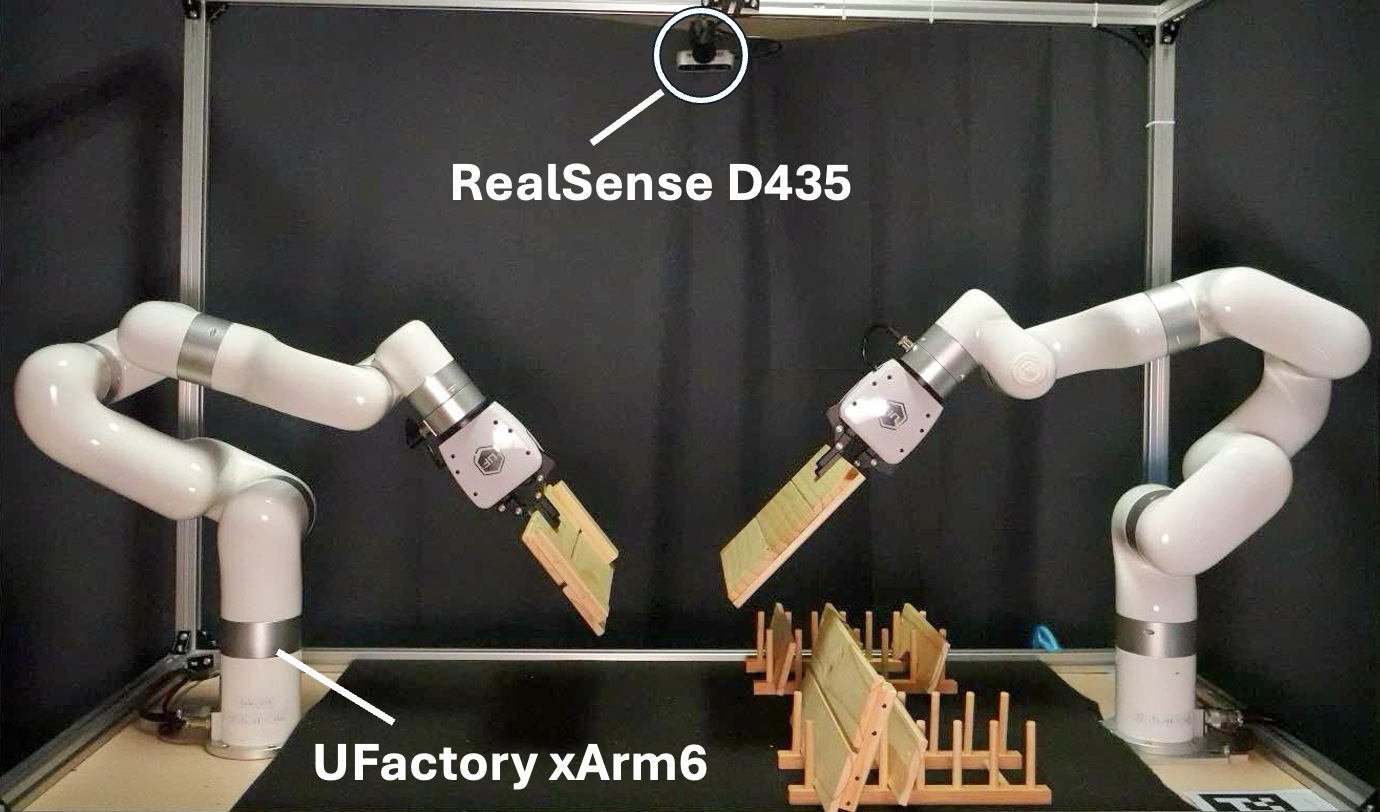}}
    \caption{\textbf{Real-World Setup.} We use two UFactory xArm6 for assembly and a RealSense D435 camera for pose estimation.}
    \label{fig:real_setup}
\end{figure}

\subsection{Real-world Assembly Experiments}~\label{real_world_exp}
To evaluate the feasibility and performance of our pipeline, we conducted experiments in the real world using four IKEA furniture items:  Flisat (Wooden Stool), Variera (Iron Shelf), Sundvik (Chair), and Knagglig (Box). \Cref{fig:real_setup} illustrates our real-world experiment setup. We show the manual images, per-step pose estimation results, and real-world assembly process in~\Cref{fig:real_assembly_process}. We also attach videos of the real-world assembly process in the supplementary material. For detailed implementation of our real-world experiments, please check~\Cref{appd: real-world}.
We evaluated all the assembly tasks with target poses provided by three different methods: Ground truth Pose, Mean-Max Pool~(see~\Cref{pose_exp}), and our proposed approach. 
The Ground truth Pose method uses the ground truth poses for each part to assemble the furniture.
We use the Average Completion Rate (ACR) as the evaluation criterion and calculate it as follows:
\begin{equation}
ACR = \frac{1}{N} \sum_{j=1}^{N} \frac{S_j}{S_{\text{total}}}
\end{equation}
where $N$ is the total number of trials, 
$S_j$ is the number of steps completed in trial $j$, and $S_{\text{total}}$ denotes the total number of steps in the task.

We perform each task over 10 trials with varying initial 3D part poses. We present the results in~\Cref{tab:real_assembly}, showing that our method outperforms the baseline and achieves a high success rate in real-world assembly tasks. 

These findings underscore the practicality and effectiveness of our approach for real-world implementation. The primary failure mode arises from planning limitations, particularly in handling complex obstacles. Failures occur when the RRT-Connect algorithm cannot find a feasible trajectory when the planned path results in collisions with the robotic arm or surrounding objects or due to suboptimal grasping poses. To improve robustness in real-world scenarios, we plan to develop a low-level policy for adaptive motion refinements—a topic we leave for future work.

\begin{table}[!ht]
    \footnotesize
    \centering
    \begin{threeparttable}
    \captionsetup{width=\linewidth}
    \caption{\textbf{Real World Success Rate ($\uparrow$) over 10 trials.}}
    \label{tab:real_assembly}
    \begin{tabular}{lccccc}
    \toprule
    {Method} & FLISAT  & VARIERA &SUNDVIK & KNAGGLIG\\ 
        \midrule
       Oracle Pose & 72.5 & 85.0 & 80.0 & 90.0 \\  
       Mean-Max Pool& 52.5 & 61.7 & 40.0& 70.0 \\ 
        Ours 
        & 60.0
        & 80.0
        & 68.0
        & 85.0 \\ 
        \bottomrule
        \end{tabular}
    
    \end{threeparttable}
    \end{table}

\subsection{Generalization to Other Assembly Tasks}
\label{gen_other_task}
\begin{figure}[htb!]
    \centering
    \includegraphics[width=0.95\linewidth]{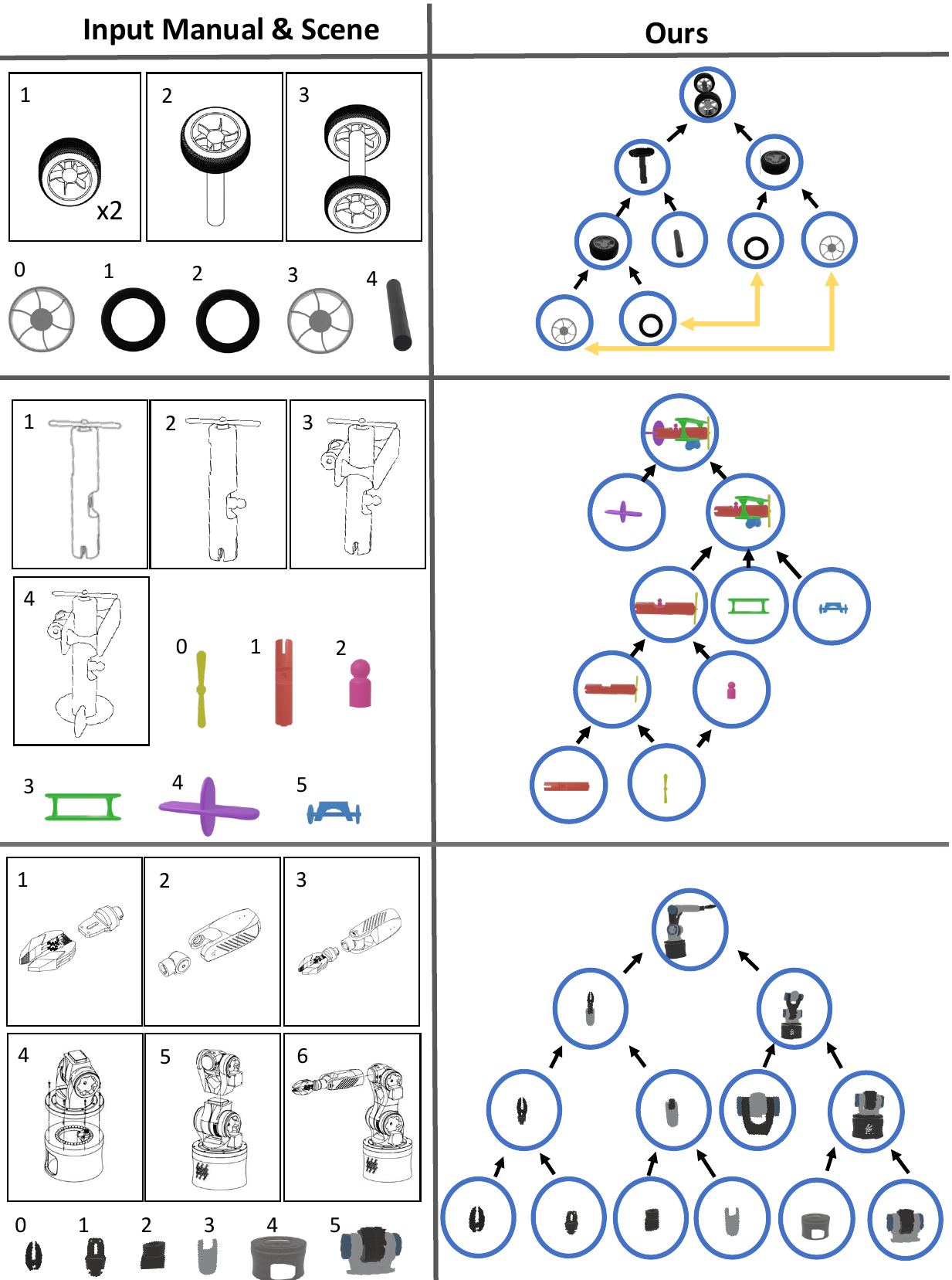}
    \caption{\textbf{Pipeline Extension Beyond Furniture Assembly.}}
    \label{fig: asap}
\end{figure}
We design Manual2Skill as a generalizable framework capable of handling diverse assembly tasks with manual instructions. To assess its versatility, we evaluate the VLM-guided hierarchical graph generation method across three distinct assembly tasks, each varying in complexity and application domain. These include: \textbf{(1) Assembling a Toy Car Axle} (a low-complexity task with standardized components, representing consumer product assembly), \textbf{(2) Assembling an Aircraft Model} (a medium-complexity task, representing consumer product assembly), and \textbf{(3) Assembling a Robotic Arm} (a high-complexity task involving non-standardized components, representing research \& prototyping assembly).

For the toy car axle and aircraft model, we sourced 3D parts from \citep{tian2024asap} and reconstructed pre-assembly scene images using Blender. We manually crafted the manuals in their signature style, with each page depicting a single assembly step through abstract illustrations. For the robotic arm assembly, we used the Zortrax robotic arm \cite{zortrax_robotic_arm}, which includes pre-existing 3D parts and a structured manual. These inputs were then processed through the VLM-guided hierarchical graph generation pipeline (described in Sec.~\ref{assembly_tree_gen}), yielding assembly graphs as shown in \Cref{fig: asap}. This \textbf{zero-shot} generalization achieves a success rate of $100\%$ over five trials per task. The generated graphs align with ground-truth assembly sequences, confirming the generalization of our VLM-guided hierarchical graph generation across diverse manual-based assembly tasks and highlighting its potential for broader applications.

\section{Limitations}
This paper explores the acquisition of complex manipulation skills from manuals and introduces a method for automated IKEA furniture assembly. Despite this progress, several limitations remain. First, our approach mainly identifies the objects that need assembly but overlooks other details on the manual, such as grasping position markings and precise connector locations (e.g., screws). Integrating a vision-language model (VLM) module to extract this information could significantly enhance robotic insertion capabilities. Second, the method does not cover the automated execution of fastening mechanisms, like screwing or insertion actions, which depend heavily on force and tactile sensing signals. We leave these challenges as directions for future work.

\section{Conclusion}

In this paper, we address the issue of learning complex manipulation skills from manuals, which is essential for robots to execute such tasks based on human-designed instructions.
We propose \ours, a novel framework that leverages VLM to understand manuals and learn robotic manipulation skills from manuals.
We design a pipeline for assembling IKEA furniture and validate its effectiveness in real scenarios.
We also demonstrate that our method extends beyond the task of furniture assembly.
This work represents a significant step toward enabling robots to learn complex manipulation skills with human-like understanding. It could potentially unlock new avenues for robots to acquire diverse complex manipulation skills from human instructions.

\bibliographystyle{plainnat}
\bibliography{references}

\clearpage
\newpage
\appendix
\lstset{
    language=Python,
    basicstyle=\ttfamily\small, 
    keywordstyle=\bfseries\color{blue}, 
    commentstyle=\itshape\color{green!50!black}, 
    stringstyle=\color{red}, 
    showstringspaces=false, 
    frame=single, 
    numbers=left, 
    numberstyle=\tiny\color{gray}, 
    breaklines=true, 
    tabsize=4, 
}


\subsection{Per-step Assembly Pose Estimation Dataset}
\label{appd: Per-step Assembly Pose Estimation Dataset}
We build a dataset for our proposed manual guided per-step assembly pose estimation task.
Each data piece is a tuple $(I_i, \{P\}_j, \{T\}_j, \mathbf{R}_i)$, where $I_i$ is the manual image, $\{P\}_j$ is the point clouds of all the components involved in the assembly step, $\{T\}_j$ is the target poses for each component, and $\mathbf{R}_i$ is the spatial and geometric relationship between components.

\begin{figure}[!ht]
    \centering
    \resizebox{0.4\textwidth}{0.22\textheight}{
    \includegraphics{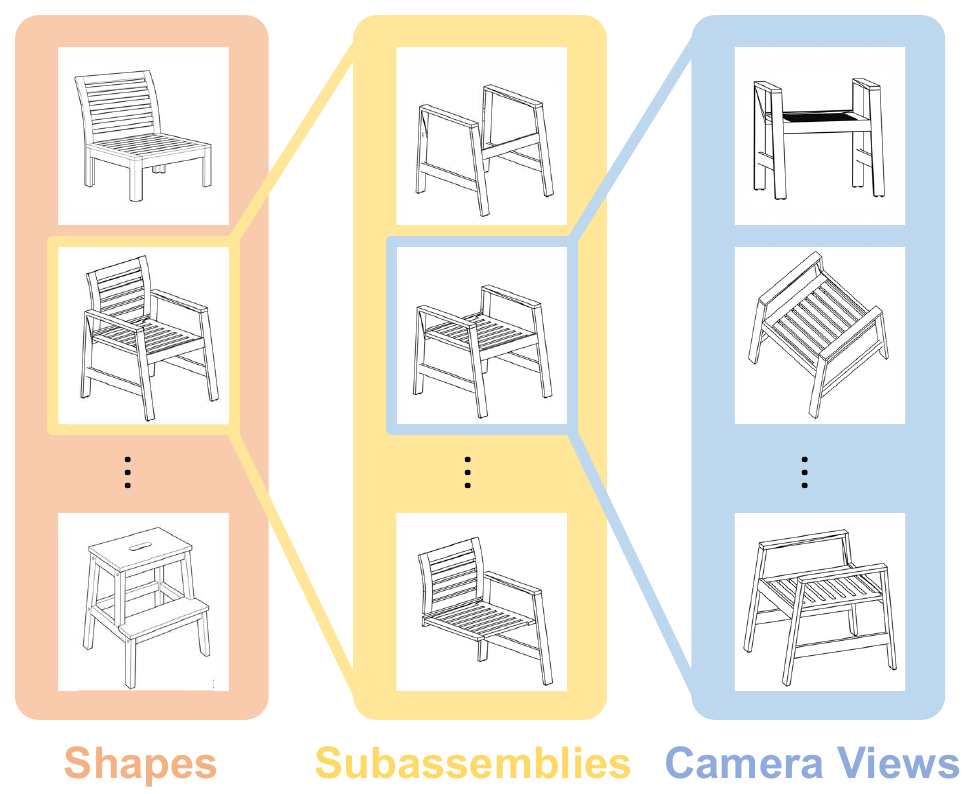}}
    \caption{Manual images of our proposed dataset. There are variations in furniture shapes, subassemblies, and camera views.}
    \label{fig: dataset}
    \end{figure}

Instruction manuals in the real world come in a wide variety. To cover as many scenarios as we might encounter in real-life situations, we considered three possible variations of instruction manuals when constructing the dataset, as shown in~\Cref{fig: dataset}.
Our dataset encompasses a variety of furniture shapes. 
For each piece of furniture, we randomly selected some connected parts to form different subassemblies. 
Meanwhile, for each subassembly, there are multiple possible camera perspectives for taking manual photos. 
This definition enables our dataset to cover various manuals that we might encounter in real-world scenarios.

Formally, for furniture consisting of $M$ parts, we randomly select $m$ connected parts to form a subassembly.
Denoted as $P_{\text{sub}} = \{P_1, P_2, \cdots, P_m\}$, here each $P_i$ is a atomic part.
Then, we randomly group the $m$ atomic parts into $n$ components while keeping all parts within the same group are connected, denoted as $P_{\text{sub}} = \{\{P_{11},\cdots P_{1\alpha_1}\},\cdots\{P_{n1},\cdots P_{n\alpha_n}\}\}$, where each $\alpha_i$ represents the number of atomic parts in $i$-th component, and thus $\sum_{i}\alpha_i=m$.
We sample the point cloud for each component to consist of the point cloud of the data piece.
We can also take photos of the subassembly from different perspectives.

We also provide annotations for equivalent parts in the auxiliary information.
In this paper, we propose new techniques to leverage the auxiliary information for each assembly step, which significantly enhances the precision and robustness of our pose estimation model.

\subsection{Pose Estimation Implementation}
\label{appd:implementation_pose}
\subsubsection{Loss Functions for Pose Estimation}  \(\) \\
\vspace{-1em}

\textbf{Rotation Geodesic Loss:}  
In 3D pose prediction tasks, we commonly use the rotation geodesic loss to measure the distance between two rotations ~\citep{wu2023leveraging}. Formally, given the ground truth rotation matrix $R \in SO(3)$ and the predicted rotation $\hat{R} \in SO(3)$, the rotation geodesic loss is defined as:
\begin{equation}
    \mathcal{L}_{\text{rot}} = \arccos\left(\frac{\text{tr}(R^T\hat{R}) - 1}{2}\right)
\end{equation}
where $\text{tr}(\cdot)$ denotes the trace of a matrix and $R^T$ is the transpose of $R$.

\textbf{Translation MSE Loss:}  
Following~\citep{li2020learning}, we use the mean squared error (MSE) loss to measure the distance between the ground truth translation $t$ and the predicted translation $\hat{t}$:
\begin{equation}
    \mathcal{L}_{\text{trans}} = ||t - \hat{t}||_2
\end{equation}

\textbf{Chamfer Distance Loss:}  
This loss function minimizes the holistic distance between each point in the predicted and ground truth point clouds. Given the ground truth point cloud $S_1 = R P + t$ and the predicted point cloud $S_2 = \hat{R} P + \hat{t}$, it is defined as:
\begin{equation}
    \mathcal{L}_{\text{cham}} = \frac{1}{|S_1|}\sum_{x \in S_1} \min_{y \in S_2} ||x - y||_{2}^{2} + \frac{1}{|S_2|}\sum_{x \in S_2} \min_{y \in S_1} ||y - x||_{2}^{2}
\end{equation}
where $S_1$ is the point cloud after applying the ground truth 6D pose transformation, and $S_2$ is the point cloud after applying the predicted 6D pose transformation.

\textbf{Pointcloud MSE Loss:}  
We supervise the predicted rotation by applying it to the point of the component and use the MSE loss to measure the distance between the rotated point and the ground truth point:
\begin{equation}
    \mathcal{L}_{\text{pc}} = ||R P - \hat{R} P||_2
\end{equation}


\textbf{Equivalent Parts:}  
Given a set of components, we might encounter geometrically equivalent parts that we must assemble in different locations. Inspired by ~\citep{zhang2024manual}, we group these geometrically equivalent components and add an extra loss term to ensure we assemble them in different locations. For each group of equivalent components, we apply the predicted transformation to the point cloud of each component and then compute the Chamfer distance (CD) between the transformed point clouds. For all pairs $(j_1, j_2)$ within the same group, we compute the Chamfer distance between the transformed point clouds $\hat{P}_{j_1}$ and $\hat{P}_{j_2}$, encouraging the distance to be large:
\begin{equation}
\label{eq:loss_equiv}
    \mathcal{L}_{\text{equiv}} = -\sum_{\text{group}}\sum_{(j_1, j_2)} \text{CD}(\hat{P}_{j_1}, \hat{P}_{j_2})
\end{equation}

Finally, we define the overall loss function as a weighted sum of the above loss terms:
\begin{equation}
    \mathcal{L}_{\text{total}} = \lambda_{1}\mathcal{L}_{\text{rot}} + \lambda_{2}\mathcal{L}_{\text{trans}} + \lambda_{3}\mathcal{L}_{\text{cham}} + \lambda_{4}\mathcal{L}_{\text{pc}} + \lambda_{5}\mathcal{L}_{\text{equiv}}
\end{equation}
where $\lambda_{1}=1$, $\lambda_{2}=1$, $\lambda_{3}=1$, $\lambda_{4}=20$,  $\lambda_{5}=0.1$. 
\vspace{1em}

\subsubsection{Mean-Max Pool}
\label{appd:implementation_mean_max_pool}
The core mechanic of the mean-max pool is to obtain the mean and maximum values along one dimension $\mathbb{R}^{C}$ of a set of vectors or matrices with the same dimensions and concatenate them into a one-dimensional vector in $\mathbb{R}^{2C}$ to obtain a global feature. For one-dimensional vectors, we take the mean and maximum values along the sequence length dimension. For two-dimensional matrices, we take the mean and maximum values along the height \(\times\) width dimensions:
\begin{equation}
    \textbf{F}_{global} = [\textbf{avg};\textbf{max}] \in \mathbb{R}^{2F}
\label{eq:mean-max pool}
\end{equation}
In the setting of our work, we set $F$ to 128.

We use this trick twice in this work. One instance is when we obtain a one-dimensional vector with a channel dimension from a multi-channel feature map, thus obtaining a one-dimensional feature vector for the image. In this case, we can express the mean-max pool as follows:
\begin{equation}
    \left\{
    \begin{aligned}
    & \textbf{X}=(\textbf{X}_{c,h,w})^{C,H,W}_{c=1,h=1,w=1}\\
    & \textbf{avg} = (\frac{1}{HW} \sum_{\textit{h}=1}^{\textit{H}} \sum_{\textit{w}=1}^{\textit{W}} \textbf{X}_{c,h,w})^{C}_{c=1} \in \mathbb{R}^{C} \\
    & \textbf{max} = (\max_{h,w} \textbf{X}_{c,h,w})^{C}_{c=1} \in \mathbb{R}^{C} \\ 
    \end{aligned}
    \right.
\end{equation}
Where $\mathbf{X}$ is the multi-channel feature map of image $I_i$ with dimensions $\text{channels} (C) \times \text{height} (H) \times \text{width} (W)$, $\mathbf{avg}$ and $\mathbf{max}$ denote one-dimensional vectors of length $\text{channels}$. Thus, $\textbf{F}_{global}$ of the multi-channel feature map is a C-dimensional vector.

The other instance is when we compare the baseline. To aggregate point cloud features on a per-part basis and obtain a one-dimensional global feature for the shape, we express the mean-max pool in the following form:
\begin{equation}
    \left\{
    \begin{aligned}
    & \textbf{avg} = \frac{1}{M}\sum_{j=1}^{M} \textbf{F}_{j} \in \mathbb{R}^{F} \\
    & \textbf{max} = \max_{F} \{\textbf{F}_{j}\} \in \mathbb{R}^{F}
    \end{aligned}
    \right.
\end{equation}
Here, we let \(M\) denote the number of parts in a shape. For each part in this baseline, we concatenate the one-dimensional image feature \(\textbf{F}_{I}\), the global point cloud feature \(\textbf{F}_{global}\) (both obtained by mean-max pool), and the part-wise point cloud feature \(\textbf{F}_{j}\) to form a one-dimensional cross-modality feature. We then use this feature as input for the pose regressor MLP.

\subsubsection{Hyperparameters in Training of Pose Estimation}
We train our pose estimation model on a single NVIDIA A100 40GB GPU with a batch size of 32. Each experiment runs for 800 epochs (approximately 46 hours). We set the learning rate to \(1e-5\) and employ a 10-epoch linear warm-up phase. Afterward, we use a cosine annealing schedule to decay the learning rate. We also set the weight decay to \(1e-7\). The optimizer configuration for each component of the model is as shown in~\Cref{tab: pose estimation optimizer}.
\begin{table}[!ht]
    \caption{Optimizer Corresponding to Each Component}
    \label{tab: pose estimation optimizer}
    \centering
    \begin{threeparttable}
    \begin{tabular}{lc}
    \toprule
    {Component} & Optimizer \\ \midrule
       Image Encoder &  RMSprop\\  
       Pointcloud Encoder&  AdamW\\ 
       GNN &  AdamW\\  
       Pose Regressor&  RMSprop\\ 
        \bottomrule
        \end{tabular}
    \end{threeparttable}
    \end{table}

\subsection{Pose Estimation Ablation Studies}
\label{appd:pose_ablation}
\begin{figure}[htb!]
    \centering
    \begin{minipage}{\linewidth}
        \centering
        \includegraphics[width=\textwidth]{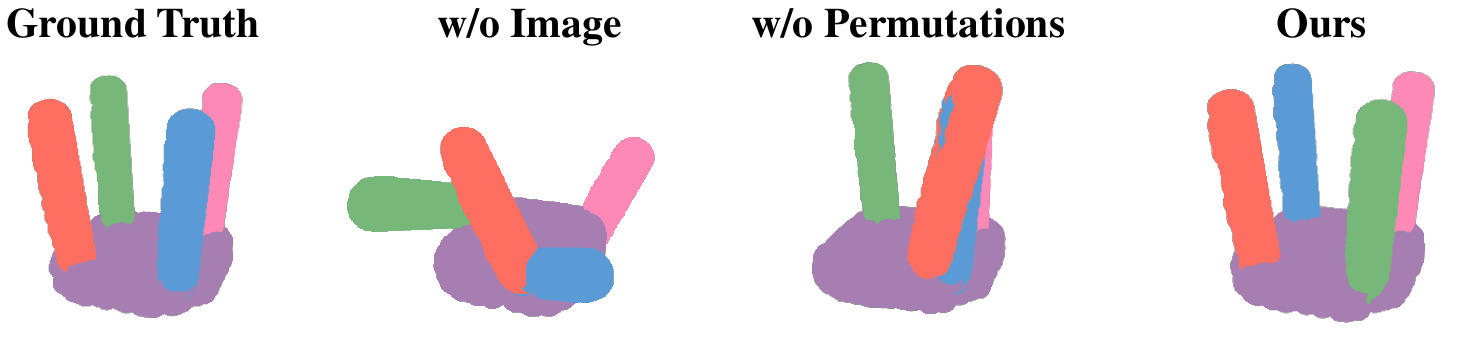}
        \caption{\textbf{Qualitative Results of Ablations.} We observe salient performance drops in ablated settings. }
        \label{fig:pose_estimation_ablation}
    \end{minipage}
\end{figure}

To evaluate the effectiveness of each component in our pipeline, we conduct an ablation study on the chair category. We show the quantitative results in~\Cref{tab:pose_estimation_ablation} and the qualitative results in~\Cref{fig:pose_estimation_ablation}.
First, we remove the image input and only use the point cloud input to predict the pose.
The performance drops significantly, indicating that the image input is crucial for pose estimation.
Second, we remove the permutation mechanism for equivalent parts(\Cref{eq:loss_equiv}).
As shown in the visualizations, the model fails to distinguish between equivalent parts, placing two legs in similar positions.

        
    

\begin{table}[htb!]
    \setlength{\tabcolsep}{4pt}
    \centering
    \begin{threeparttable}
    \caption{\textbf{Pose Estimation Ablations.}}
    \label{tab:pose_estimation_ablation}
    \begin{tabular}{lccccc}
    \toprule 
    {Method} & GD$\downarrow$  & RMSE$\downarrow$ &CD$\downarrow$ & PA$\uparrow$ \\ 
        \midrule
        w/o Image&1.797  &0.234  &0.227  &0.138 \\
        w/o Permutations&0.252  &0.051  &0.029  &0.783 \\

        \textbf{Ours}
        &\cellcolor{customblue}\textbf{0.202}  
        &\cellcolor{customblue}\textbf{0.042}  
        &\cellcolor{customblue}\textbf{0.027} 
        &\cellcolor{customblue}\textbf{0.868} \\ 
        
        \bottomrule
        \end{tabular}
    
    \end{threeparttable}
\end{table}

Previous works usually train and predict only fully assembled shapes. In contrast, our pose estimation dataset includes per-step data ($i.e.$, subassemblies). We conduct an ablation study comparing two settings:
\begin{itemize}
    \item \textit{w/o Per-step}: Training and testing on a dataset of fully assembled shapes.
    \item \textit{Per-step}: Training on a dataset with per-step data and testing on fully assembled shapes.
\end{itemize}

\begin{table}[!ht]
    \setlength{\tabcolsep}{7pt}
    \centering
    \begin{threeparttable}
    \caption{\textbf{w/o Per-step vs. Per-step}}
    \label{tab:one_step}
    \begin{tabular}{lccccc}
    \toprule 
    {Method} & GD$\downarrow$  & RMSE$\downarrow$ &CD$\downarrow$ & PA$\uparrow$\\ 
        \midrule
        w/o Per-step &0.233  &0.046  &0.015 &0.753  \\
        \cellcolor{customblue}Per-step\textbf{ (Ours)}
        &\cellcolor{customblue}\textbf{0.064}  
        &\cellcolor{customblue}\textbf{0.016}  
        &\cellcolor{customblue}\textbf{0.004} 
        &\cellcolor{customblue}\textbf{0.983}\\ 
        \bottomrule
        \end{tabular}
    
    \end{threeparttable}
    
\end{table}

As shown in~\Cref{tab:one_step}, adding per-step data improves assembly prediction accuracy, demonstrating that per-step inference enhances robot assembly performance.

\subsection{Additional VLM Plan Generation Results}
\label{vlm_plan_complete}
Besides using the Success Rate metric defined in \Cref{assembly_tree_gen} for evaluating VLM assembly graph generation, we also provide additional analysis using evaluation metrics in the IKEA-Manuals dataset \cite{wang2019dynamic}. These include Precision, Recall, F1 Score, Simple Matching, and Hard Matching as seen in \Cref{tab:full_asssembly_plan}. For detailed descriptions of these metrics, we refer
readers to \cite{wang2022ikea}.
Furthermore, we categorized the full set of 102 furniture items in greater detail, with Hard Matching results for individual part counts ranging from 2 to 16 parts, as shown in \Cref{tab:parts_performance}.


To obtain our scores, we repeatedly ran the experiment 7 times using the same input and a temperature of 0. We repeated sampling to account for slight variations in GPT-4o's \cite{achiam2023gpt} outputs, even when we set the temperature to 0, and to capture the range of possible outcomes. This approach provides a better estimate of the model's true performance. We choose to report the best score in all of our tables related to Assembly Plan Generation, since the average and best scores are similar and within a difference of 5\% in all tests.

To compare the trees generated by GPT-4o \cite{achiam2023gpt} with the ground truth trees in the dataset, we accounted for equivalence relationships among parts, which can result in multiple valid ground truth trees. For instance, if parts 1 and 2 are equivalent and [[1, 3], 2] is a valid tree, then so is [[2, 3], 1]. Since the dataset does not account for this isomorphism of trees, we manually defined all equivalent parts for each of the 102 furniture items. We then permuted the predicted tree using the equivalent parts, comparing each permutation to the ground truth and selecting the highest score. For furniture with 13 or more parts (6 items), we performed manual verification due to the computational cost of permutations. Overall, by employing this permutation method to evaluate predicted trees, we managed to increase our scores overall metrics by around 5\%. To ensure fairness, we also applied this permutation over the two baselines but saw no effects.

As noted in \cite{wang2022ikea}, SingleStep always outputs the root node and selects all other nodes as its children, achieving perfect precision in Simple Matching for all cases. Beyond this, our GPT-4o-based method outperforms both baselines across all categories in \Cref{tab:full_asssembly_plan}, which highlights the effectiveness of VLMs in interpreting manuals and designing reliable hierarchical assembly graphs.

Similarly, in \Cref{tab:parts_performance}, our method has a significant advantage over the two baselines in all numbers of parts. The performance score shown is calculated by taking the maximum of Precision, Recall, and F1 Score. Mask Seg is an additional method we evaluated, which overlays segmentation masks from the IKEA-Manuals dataset \cite{wang2022ikea} onto manual pages (prompt 3.a~\Cref{appd: additional_prompts}), improving part identification, image clarity, and comprehension of assembly steps. Although Mask Seg slightly outperforms the original version without mask segmentations, we chose the latter for all reported tables. Otherwise, such masks are costly in real-world scenarios. Overall, the trend indicates with higher scores and accuracies for furniture with fewer parts and lower scores as the number of parts increases.

\begin{table}[ht]
\centering
\begin{threeparttable}
\caption{VLM Assembly Plan Generation Results ($\uparrow$)}
\label{tab:full_asssembly_plan}
\renewcommand{\arraystretch}{1.2}
\begin{tabular}{@{}lccc|ccc@{}}
\toprule
\textbf{Method} & \multicolumn{3}{c|}{\textbf{Simple Matching}} & \multicolumn{3}{c}{\textbf{Hard Matching}} \\
\cmidrule(lr){2-4} \cmidrule(lr){5-7}
& Precision & Recall & F1 & Precision & Recall & F1 \\
\midrule
SingleStep & \textbf{100.00} & 35.77 & 48.64 & 10.78 & 10.78 & 10.78 \\
GeoCluster & 44.90 & 48.46 & 43.53 & 16.54 & 16.50 & 16.30 \\
\rowcolor{customblue}
\textbf{Ours} & 58.11 & \textbf{55.98} & \textbf{56.84} & \textbf{40.63} & \textbf{39.94} & \textbf{40.22} \\
\bottomrule
\end{tabular}
\end{threeparttable}
\end{table}

\begin{table*}[ht]
    \centering
    \setlength\tabcolsep{1pt} 
    \renewcommand{\arraystretch}{1.2} 
    \begin{threeparttable}
    \caption{Performance Across Different Numbers of Parts  ($\uparrow$)}
    \label{tab:parts_performance}
    \begin{tabular}{@{}l>{\centering\arraybackslash}p{1cm} >{\centering\arraybackslash}p{1cm} >{\centering\arraybackslash}p{1cm} >{\centering\arraybackslash}p{1cm} >{\centering\arraybackslash}p{1cm} >{\centering\arraybackslash}p{1cm} >{\centering\arraybackslash}p{1cm} >{\centering\arraybackslash}p{1cm} >{\centering\arraybackslash}p{1cm} >{\centering\arraybackslash}p{1cm} >{\centering\arraybackslash}p{1cm} >{\centering\arraybackslash}p{1cm} >{\centering\arraybackslash}p{1cm} >{\centering\arraybackslash}p{1cm} >{\centering\arraybackslash}p{1cm}@{}}
    \toprule
    \textbf{Number of Parts} & \textbf{2} & \textbf{3} & \textbf{4} & \textbf{5} & \textbf{6} & \textbf{7} & \textbf{8} & \textbf{9} & \textbf{10} & \textbf{11} & \textbf{12} & \textbf{13} & \textbf{14} & \textbf{15} & \textbf{16} \\ 
    \midrule
    SingleStep & \textbf{100} & 50 & 12.50 & 31.58 & 0 & 0 & 0 & 0 & 0 & 0 & 0 & 0 & 0 & 0 & 0 \\ 
    GeoCluster & \textbf{100} & 25 & 10.42 & 14.04 & 21.76 & 14.40 & 6.99 & 15.00 & 4.17 & 2.22 & 0 & 16.67 & 0 & 0 & 0 \\ 
    Ours (Mask Seg) & \textbf{100} & \textbf{100} & \textbf{75.00} & 72.81 & \textbf{56.08} & \textbf{29.64} & \textbf{24.17} & \textbf{19.05} & 16.67 & \textbf{9.63} & 3.33 & \textbf{37.50} & 0 & 0 & 0 \\   
    Ours & \bluebold{100} & \bluebold{100} & \cellcolor{customblue}{72.92} & \bluebold{78.51} & \cellcolor{customblue}{45.59} & \cellcolor{customblue}{25.24} & \cellcolor{customblue}{13.05} & \cellcolor{customblue}{16.67} & \bluebold{27.78} & \cellcolor{customblue}{0} & \bluebold{9.33} & \cellcolor{customblue}{6.25} & \cellcolor{customblue}{0} & \cellcolor{customblue}{0} & \cellcolor{customblue}{0}\\ 
    \midrule
    Furniture Count & 2 & 4 & 8 & 19 & 17 & 14 & 10 & 3 & 4 & 9 & 5 & 2 & 1 & 2 & 1\\ 
    \bottomrule
    \end{tabular}
    \end{threeparttable}
\end{table*}

\subsection{Assembly Graph Generation Ablation Studies}
\label{appd:VLM_ablation}
Since manual pages provide critical visual details, such as parts and subassemblies for each step, that directly influence GPT-4o’s output, we evaluate two key design choices in our VLM-based plan generation pipeline: (1) The inclusion of manual pages to guide the VLM’s reasoning over assembly graphs, and (2) the use of segmented manual pages (vs. full manuals) as input.

\subsubsection{Excluding Manual Pages Entirely}
To explore whether a modified framework could perform the task without manual images by relying on the VLM’s existing priors, we ablate our pipeline by removing manual images and providing only the pre-assembly scene along with a simplified text prompt. When tested on all 102 furniture items, this variant performs poorly as shown in \Cref{tab:assembly_plan_ablation} (Ours (w/o Manual)). The VLM appears to leverage some structural priors, which produces some plausible plans for simpler items with four or fewer parts. However, it fails to generalize to more complex furniture. Manually inspecting each assembly plan reveals common failure modes: the VLM frequently misidentifies parts (e.g., labeling a bench seat as a "tabletop"), generates physically implausible sequences (e.g., attaching two chair legs without a connecting seat or support bar), and often includes only one or two parts per step, even when multiple parts should be assembled simultaneously in more complex furniture. The issues of hallucinations in part identification and logical inconsistencies in assembly steps highlight the limitations of relying solely on the VLM’s learned priors, and underscore the importance of manual guidance for generating accurate and coherent assembly plans in more complex scenarios.

\subsubsection{Using Non-Segmented Manual Pages}
Instead of providing $N$ segmented manual pages (one per assembly step), we input the entire manual consisting of $M \geq N$ pages, which introduces additional elements besides parts and subassemblies that the VLM must understand, such as nails, connectors, fasteners, and thought bubbles. As shown in \Cref{tab:assembly_plan_ablation} (Ours (w/o Segmentation)), this method significantly impair VLM performance in generating assembly graphs, leading to more than double accuracy drops in success rate. This performance drop is consistently observed across GPT-4o, o1, Claude 3.7 Sonnet, and Qwen 2.5 Max, suggesting a common limitation in current VLMs. We hypothesize that this additional information, while relevant for low-level
assembly actions, may not be useful for high-level planning. They may introduce noise for high-level planning, distracting the VLM from the core parts needed at each step. 

Overall, \Cref{tab:assembly_plan_ablation} underscores the importance of both including and segmenting manual pages to ensure accurate and reliable assembly graph generation.


\begin{table}[htb!]
    \centering
    \setlength\tabcolsep{4pt} 
    \renewcommand{\arraystretch}{1.2} 
    \begin{threeparttable}
    \captionsetup{width=\linewidth}
    \caption{\textbf{Assembly Plan Generation Ablation Results} ($\uparrow$)}
    \label{tab:assembly_plan_ablation}
    \begin{tabular}{@{}lcccccc@{}} 
    \toprule
    Number of Parts & 2$\sim$4 & 5$\sim$6 & 7$\sim$8 & 9+ & Average \\ 
    \midrule
    Ours (w/o Manuals) & 36.4 & 0 & 0 & 0 & 3.9\\ 
    Ours (w/o Segmentation) & 50.0 & 19.4 & \textbf{8.3} & 0 & 15.6 \\ 
    \textbf{Ours} & \textbf{78.6} & \textbf{55.6} & \textbf{8.3} & 0 & \textbf{32.4}\\ 
    \midrule
    Furniture Count & 14 & 36 & 24 & 28  \\ 
    \bottomrule
    \label{tab:assembly_plan}
    \end{tabular}
    \end{threeparttable}
    \vspace{-2mm}
\end{table}

\subsection{Failure Cases Analysis}
\label{appd: failure cases}
\begin{figure}[!ht]
    \centering
    \includegraphics[width=0.9\linewidth]{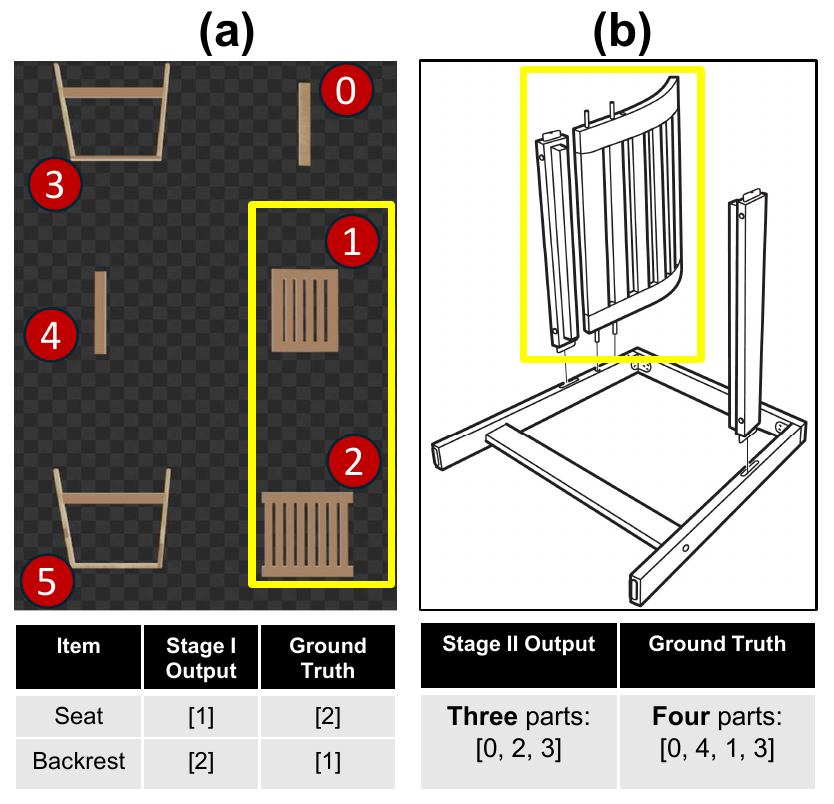}
    \caption{Two common failure patterns: (a) The VLM misidentifies parts labeled 1 and 2. (b) The VLM treats the two parts highlighted in yellow as one part, producing an incorrect assembly plan}
    \label{fig:failed-cases}
\end{figure}

\begin{figure*}[ht]
\centering
\includegraphics[page=2, width=\textwidth]{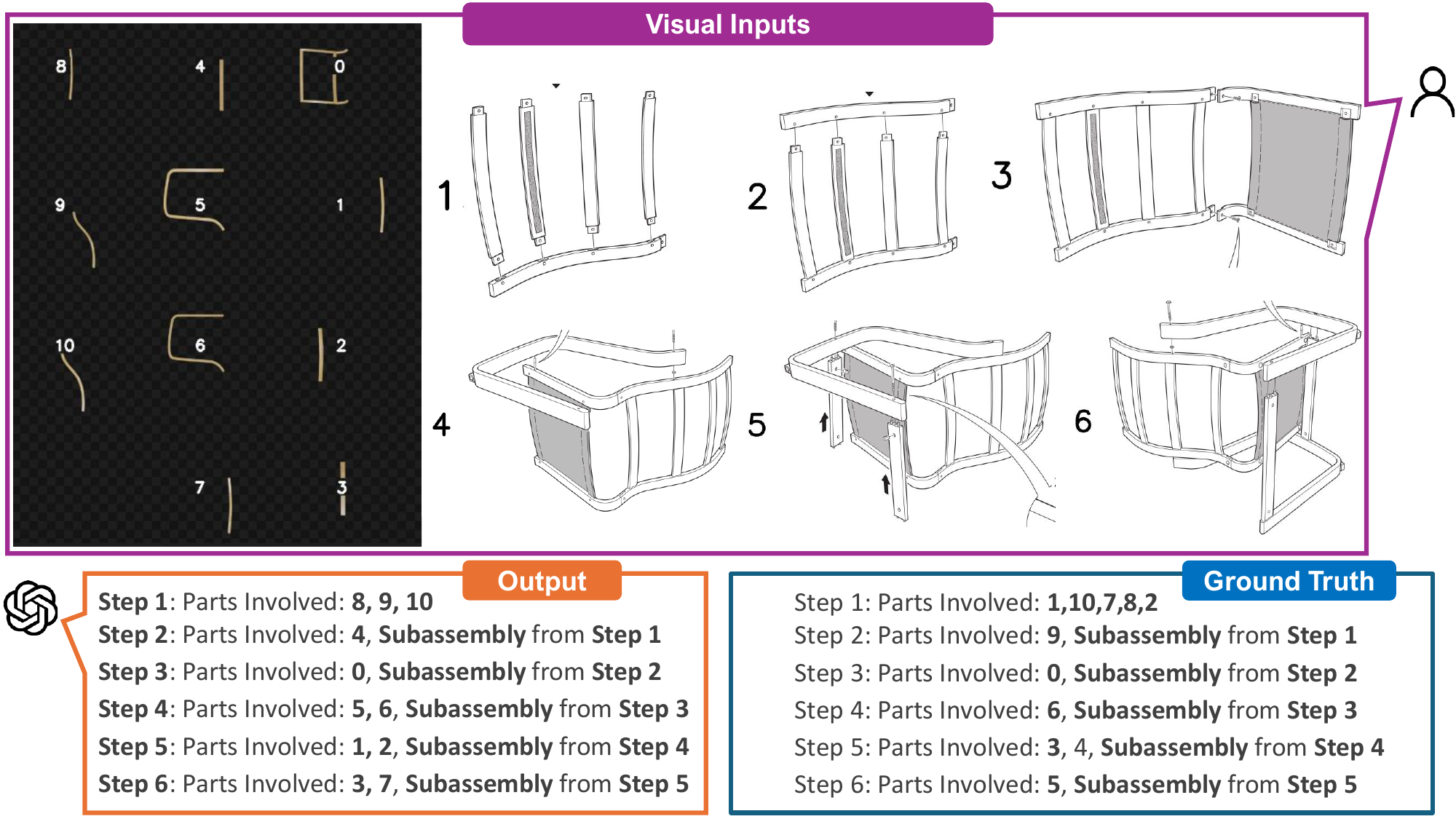} 
\caption{The input consists of the scene image, the corresponding assembly steps from the manual, and the text instruction from prompt 3.b). Clearly, GPT-4o's response is wrong and unreliable.}
\label{fig:failure2}
\end{figure*}

We analyze the failure cases in assembly graph generation. The most frequent
failure modes include: (1) The VLM misidentifies visually
similar components in the pre-assembly scene. For example,
in \Cref{fig:failed-cases}(a), the backrest and seat are confused due to similar
shapes. (2) The VLM struggles to correctly interpret parts in
the manual when they are presented in uncommon orientations,
clustered closely together, or partially occluded. For example,
in \Cref{fig:failed-cases}(b), the two parts marked by yellow box are too close in
the manual image, causing the VLM to recognize them as single component.
\Cref{fig:failure2} further demonstrates that the VLM struggles with complex structures, often producing entirely incorrect results, due to a combination of failure cases mentioned above.

\subsection{Real-World Experiment Details}
\label{appd: real-world}

This section provides the details of the real-world experiment.

\subsubsection{Pose Estimation in the Real World}

We utilize FoundationPose~\cite{wen2024foundationpose} to evaluate the 6D pose and point cloud of components in the real-world scene. First, a mobile app, ARCode, is used to scan the mesh of all atomic parts of the furniture. During each step of the assembly process, the mesh—along with the RGB and depth images and an object mask—is input into the FoundationPose model, which then generates the precise 6D pose and point cloud of the component within the scene. This information is crucial for subsequent tasks, including camera pose alignment, grasping, and collision-free planning.

\subsubsection{Camera Frame Alignment}
After we get the estimated target pose, we first use the PCA mentioned before to canonize them. To accurately map these target poses to the real world, we need to align the camera frame in the manual page image, denoted as \(P_{m_i}\), with the real-world camera frame, denoted as \(P_{w_i}\), for each step \(i\). This section will introduce how we calculate the 6D transformation matrix \(T_{mw}\) between these two frames.

To achieve this, we designate a stable part of the scene as a base in the world frame using the VLM and utilize FoundationPose to extract the point cloud of this part. We then canonicalize the point cloud using the same PCA algorithm, ensuring that the relative 6D pose of the same component remains consistent. We denote the canonical base pose in the real world as \(P_{B_w}\), which remains static during this step. From the model's predictions, we can also determine the pose of the same part used as the base in the manual, denoted as \(P_{B_m}\). We denote the transformation matrix between these two frames as \(T_{mw}\). Using this transformation matrix, we map the target pose in the manual frame, \(P_{T_m}\), to the corresponding target pose in the real-world frame, \(P_{T_w}\), for subsequent motion planning. We compute the transformation as follows:

\[
T_{mw} = P_{B_w} P_{B_m}^{-1}
\]

We then calculate the target pose in the real-world frame using:

\[
P_{T_w} = T_{mw} P_{T_m}
\]

\begin{figure}[!ht]
    \centering
    \resizebox{0.48\textwidth}{0.22\textheight}{
    \includegraphics{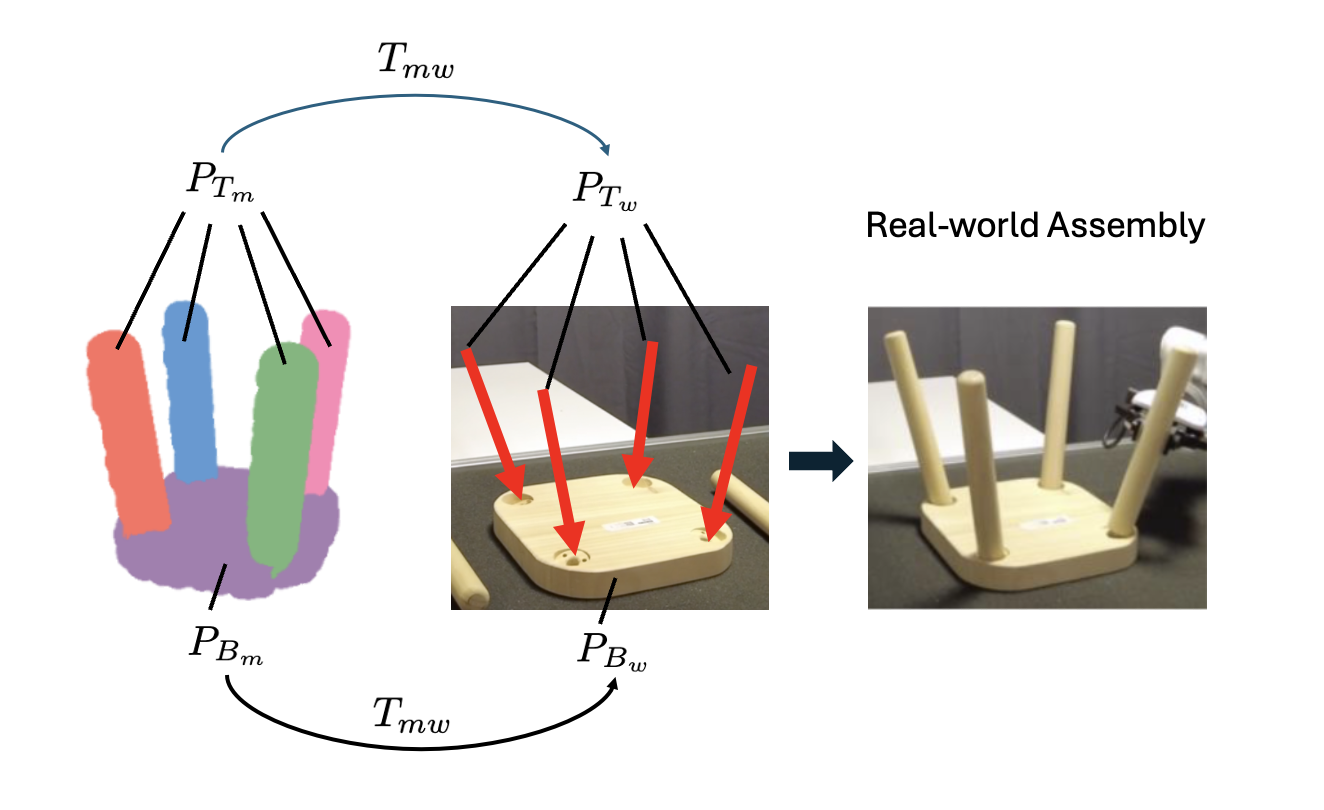}}
    \caption{This figure shows the transformation between the estimated pose and the real-world frame; we designate the board of the stool as a base and map the four legs of the stool to the real world}
    \label{fig:appendix-alignment}
\end{figure}

As illustrated in ~\Cref{fig:appendix-alignment}, the stool example clearly demonstrates the process of aligning poses between the manual and real-world frames, ensuring a consistent and reliable foundation for motion planning.

\subsubsection{Heuristic Grasping Policy}  
For general grasping tasks, pre-trained models such as GraspNet\cite{fang2023anygrasp} are commonly used to generate grasping poses. However, in the case of furniture assembly, where components are often large and flat, we need to grasp specific parts of the object that are suitable for subsequent assembly. This requirement poses challenges for GraspNet, as it does not always estimate the best pose for the subsequent action. To address this, in addition to GraspNet, we utilize the poses generated by FoundationPose and consider the shapes of the furniture components in corner cases. These shapes are categorized into two types, as shown in ~\Cref{fig:appendix-grasping}:
\begin{figure}[!ht]
    \centering
    \resizebox{0.48\textwidth}{0.22\textheight}{
    \includegraphics{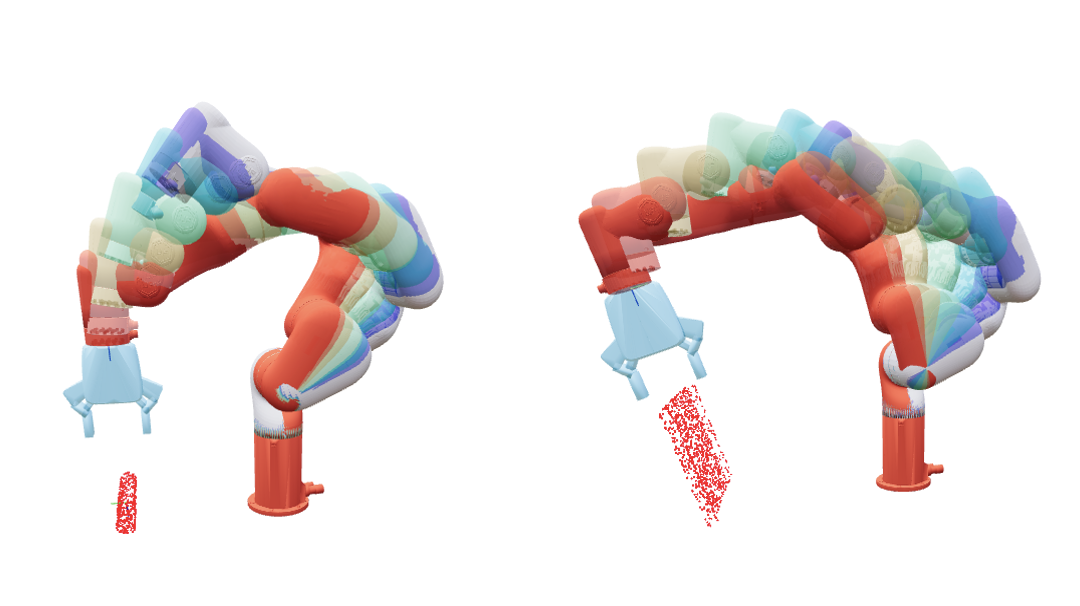}}
    \caption{This figure shows the grasping policies for different shapes in our setting; the left one is for stick-shaped, and the right one is for flat, thin-shaped.}
    \label{fig:appendix-grasping}
\end{figure}

\textbf{Stick-Shaped Components}:  
For stick-shaped furniture parts, such as stool legs, we select the center of the point cloud as the grasping position. We define the grasping pose as a top-down approach.

\textbf{Flat and thin-Shaped Components}:  
We first estimate the pose of flat and thin, board-shaped furniture parts using a bounding box. Based on this estimate, we determine the grasping pose by aligning it with the bounding box's orientation. The grasping position is set approximately 3 cm below the top surface.

\subsection{Rationale for Excluding Performance Evaluation of Stage I in Hierarchical Assembly Graph Generation}
\label{appd: no_stage1}

Stage I, Associating Real Parts with Manuals, focuses on associating real parts with the manual. Still, since the IKEA manual lacks isolated images of individual parts, direct quantitative evaluation is challenging. Instead, Stage II implicitly reflects the quality of these associations by outputting the indices of identified real parts. Therefore, we report Stage II results as an intermediate measure of how effectively our approach aligns manual images with real components.

\subsection{Justification for Hierarchical Assembly Graph}
\label{def_assembly_tree}

Using a hierarchical structure to represent assembly steps provides several advantages over simple linear data structures or unstructured step-by-step plans in plain text.

\begin{itemize}
\item Hierarchical structures align naturally with the assembly process where multiple parts and subassemblies combine into larger subassemblies.
\item Lists or text plans struggle to store geometric and spatial relationships between each part or subassembly of the step, which is crucial in real assembly tasks.


\item The hierarchical graph clearly shows the dependencies between steps, revealing which steps you can perform in parallel and which ones you must complete before proceeding to others. 
So, it provides flexibility for parallel construction or strategic sequencing.
\end{itemize}

\subsection{Formal Definition of Hierachial Assembly Graph}
\label{appd:def_assembly_graph}

Inspired by~\citet{mo2019structurenet}, we represent the assembly process as a hierarchical graph $S=(\mathbf{P,H,R})$.
A set of nodes $\mathbf{P}$ represents the parts or subassemblies in the assembly process. 
A structure $(\mathbf{H, R})$ describes how these nodes are assembled and related to each other.
The structure consists of two edge sets: $\mathbf{H}$ describes the \textit{assembly relationship} between nodes, and $\mathbf{R}$ represents the \textit{geometric} and \textit{spatial} relationship between nodes.

\textbf{Node}.
Each node $v\in\mathbf{P}$ is an atomic part or a subassembly, consisting of a non-empty subset of parts $p(v)\subset\mathcal{P}$.
The root node \( v_N \) represents the fully assembled furniture, with \( p(v_N) = \mathcal{P} \).
A non-root, non-leaf node \( v_i \) represents a subassembly with \( p(v_i) \) as a non-empty and proper subset of \( \mathcal{P} \).
All leaf nodes \( v_l \) represent atomic parts, containing exactly one element from \( \mathcal{P} \).
Additionally, each non-leaf node corresponds to a manual image \( I \) that describes how to merge smaller parts and subassemblies to form the node.

\textbf{Assembly relationship}.
We formulate the assembly process as a tree, with all atomic parts serving as leaf nodes. The atomic parts are then recursively combined into subassemblies, forming non-leaf nodes until they reach the root node, which represents the fully assembled furniture. The directed edges from a child node to its parent node indicate the assembly relationship.
The edge set \( \mathbf{H} \) includes directed edges from a child node to its parent node, indicating the assembly relationship.
For a non-leaf node $v_i$, denote its child nodes as $C_i$, the following properties hold:
\begin{enumerate}[label=(\alph*)]
    \item $\forall v_j \in C_i, p(v_j)$ is a non-empty subset of $\mathcal{P}$
    \item All children nodes contain distinct elements
    \begin{equation}
        p(v_j) \cap p(v_k) = \emptyset, \forall v_j, v_k \in C_i, j\neq k
    \end{equation}
    \item The union of all child subsets equals \( p(v_i) \):  
    \begin{equation}
        \bigcup_{v_j\in C_i} p(v_{j}) = p(v_i)
    \end{equation}
    
\end{enumerate}

\textbf{Equivalence relationship}.
In addition to the assembly process's hierarchical decomposition, we also consider the equivalence relationship between nodes. 
We label two parts \textit{equivalent} if they share a similar shape and can be used interchangeably in the assembly process.
We represent this relationship with undirected edges $\mathbf{R_i}$ in child nodes $C_i$ of node $v_i$.
An edge $\{v_a,v_b\}\in R_i$ appears between two nodes \(v_a \in C_i\), \(v_b \in \mathcal{P}\), if the shape represented by \(v_a\) and \(v_b\) are geometric equivalent and thus can be changed during assembly. 
Note that \(v_b\) is not constrained as a child of $v_i$ since any two nodes could be equivalent, regardless of their hierarchical positions.

The assembly structure is a hierarchical graph, where the nodes represent parts or subassemblies, and the edges represent the assembly and equivalence relationships.
We consider this structured representation to be a more informative and interpretable way to formulate the assembly process than a flat list of parts.

\subsection{Prompts}
\label{appd: additional_prompts}
We offer a comprehensive set of prompts utilized in the VLM-guided hierarchical graph generation process. The process involves four distinct prompts, divided into two stages. The first two prompts, which are slight variations of each other, are used in \emph{Stage I: Associating Real Parts with Manuals}. The remaining two prompts, also slight variations of each other, are employed in \emph{Stage II: Identifying Involved Parts in Each Step}.
\begin{enumerate}
\item The first prompt is part of Stage I, and it initializes the JSON file's structure and consists of two sections:
\begin{itemize}
    \item 1.a): \textbf{Image Set:} An image of the scene with furniture parts labeled using GroundingDINO \cite{liu2025grounding}, alongside an image of the corresponding manual's front page.
    \item 1.b) \textbf{Text Instructions:} A few sentences explaining the JSON file generation, supported by an example of the desired structure via in-context learning.
\end{itemize}
This prompt is passed into GPT-4o to generate a JSON file with the name and label for each part. 

\item The second prompt belongs in Stage I as well, and it populates the JSON file with detailed descriptions of roles. It includes:
\begin{itemize}
    \item 2.a): \textbf{Image Set:} Images of all manual pages (replacing the front page) to provide context about the function of each part and the scene image from the first prompt.
    \item 2.b): \textbf{Text Instructions:} a simple text instruction explaining the context and output.
\end{itemize}
We combine the JSON output from the first prompt with the second prompt, then query GPT-4o to generate the populated JSON file.

\item The third prompt is a part of Section II, and it generates a step-by-step assembly plan using:
\begin{itemize}
    \item 3.a): \textbf{Image Set:} The scene image and cropped manual pages highlight relevant parts and subassemblies, helping GPT-4o focus on key details. The cropped images also have a highlighted black number on the left, indicating the current assembly step. Our ablation studies demonstrate the effectiveness of these cropped images.
    \item 3.b): \textbf{Text Instructions:} A text instruction combining chain-of-thought and in-context learning to describe the assembly plan generation process and guide the VLM.
The JSON file from Step 2 is concatenated with the third prompt as input, guiding GPT-4o to produce the final text-based assembly plan.
\end{itemize}

\item Section II includes the fourth prompt, which converts the text-based plan into a traversable tree structure for action sequencing in robotic assembly. We achieve this conversion using a simple text input with in-context learning examples.
\end{enumerate}

\twocolumn[{
    \begin{tcolorbox}[title=1.a) \textbf{Image Set} for JSON File Generation]
    \label{JSON_1a}
        \begin{center}
            \includegraphics[width=0.3\textwidth]{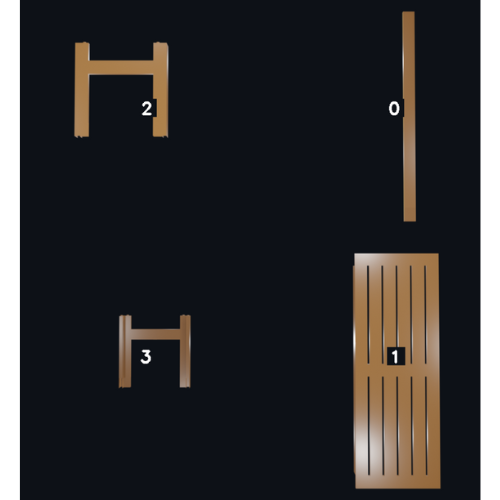}
            \includegraphics[width=0.3\textwidth]{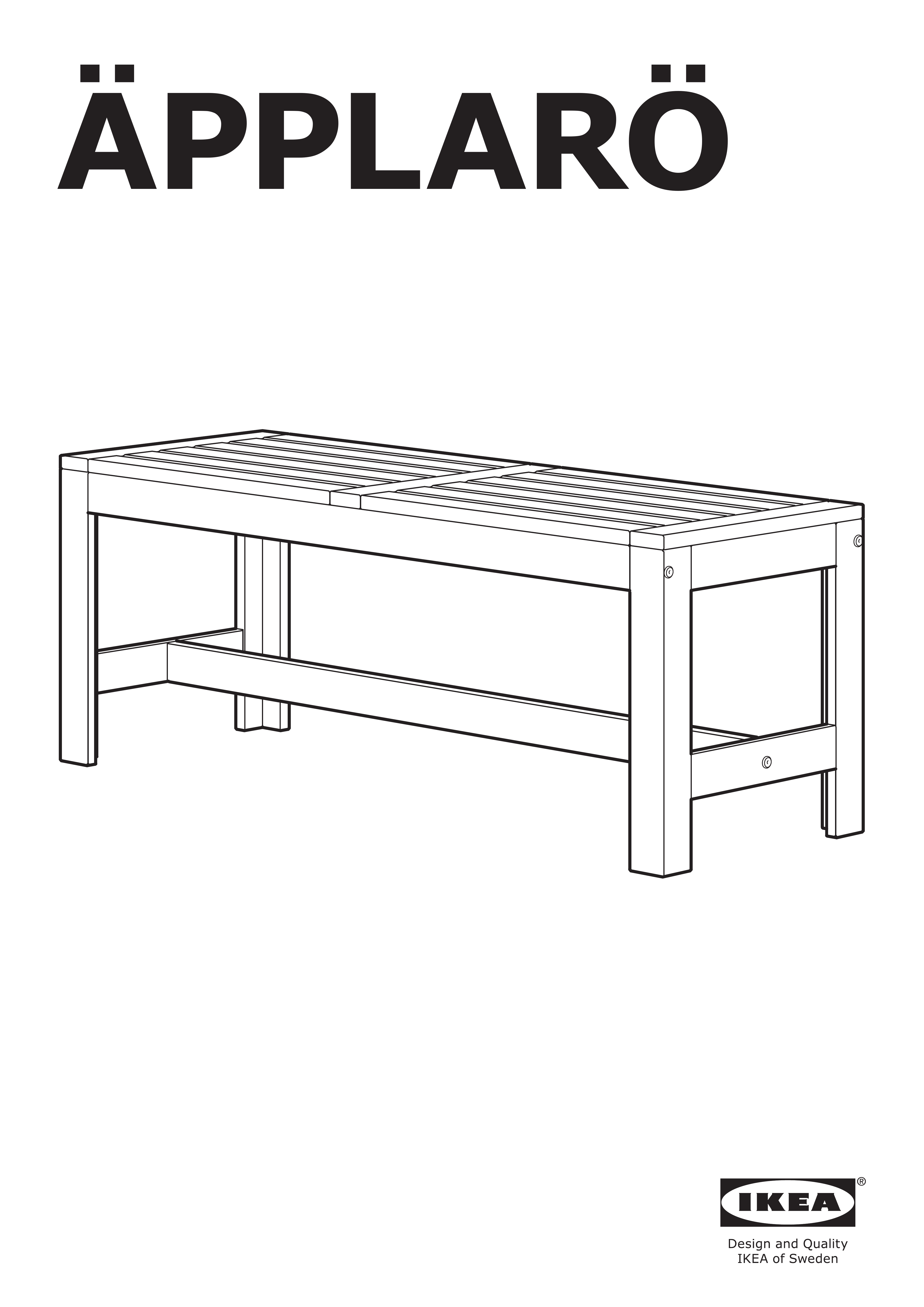}
        \end{center}\textbf{}
        \vspace{0.5em}
    \end{tcolorbox}

\begin{tcolorbox}[title=1.b) \textbf{Text Instructions} for JSON File Generation]
Input is one image, which is a top view of all the parts of one piece of furniture, each has a number, and another image, which is the first page of the setup manual\newline

You should list all the parts in the image, determine their number and name (short description of the part), and show your result in JSON format.\newline

Following is an example. Note that your output should only contain the JSON code without any explanation.\newline

\#\#\#\#\#\#\#\#\#\# example start \#\#\#\#\#\#\#\#\#\#


[\\
    \{ \\
      "name": "seat frame," \\
      "number": [0] \\
    \}, \\
    \{\newline
      "name": "side leg," \newline
      "number": [1]\newline
    \},\newline
    \{\newline
      "name": "side le," \newline
      "number": [2]\newline
    \},\newline
    \{\newline
        "name": "support b," "\newline
        "number": [3]\newline
    \}\newline
]\newline
\#\#\#\#\#\#\#\#\#\# example end \#\#\#\#\#\#\#\#\#\#
\end{tcolorbox}
}]

\twocolumn[{
    \begin{tcolorbox}[title=2.a) \textbf{Image Set} for JSON File Refinement]
        \begin{center}
            \includegraphics[width=0.22\textwidth]{Figures/scene_annotated.png}
            \includegraphics[width=0.22\textwidth]{Figures/page_1.png}
            \includegraphics[width=0.22\textwidth]{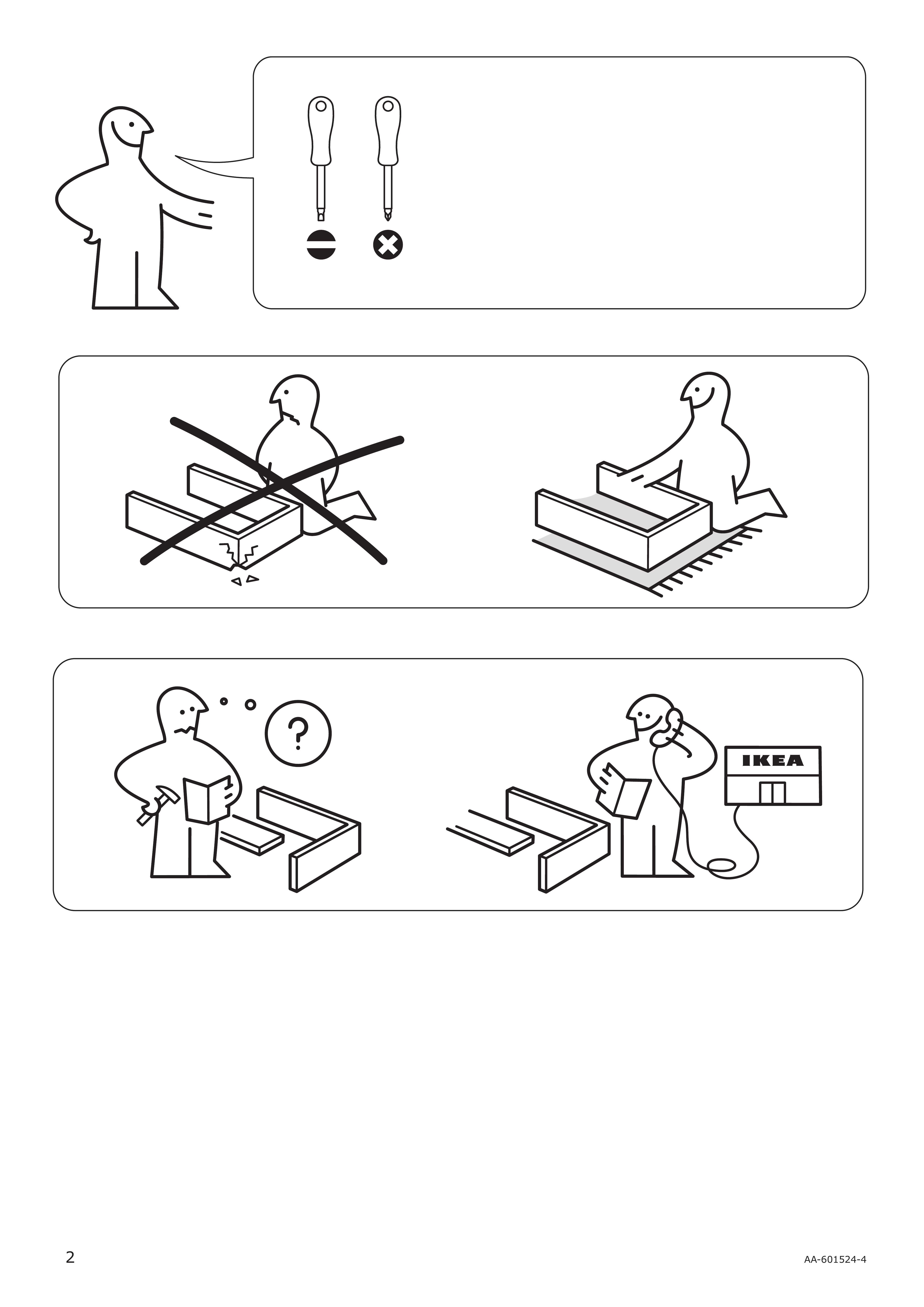}
            \includegraphics[width=0.22\textwidth]{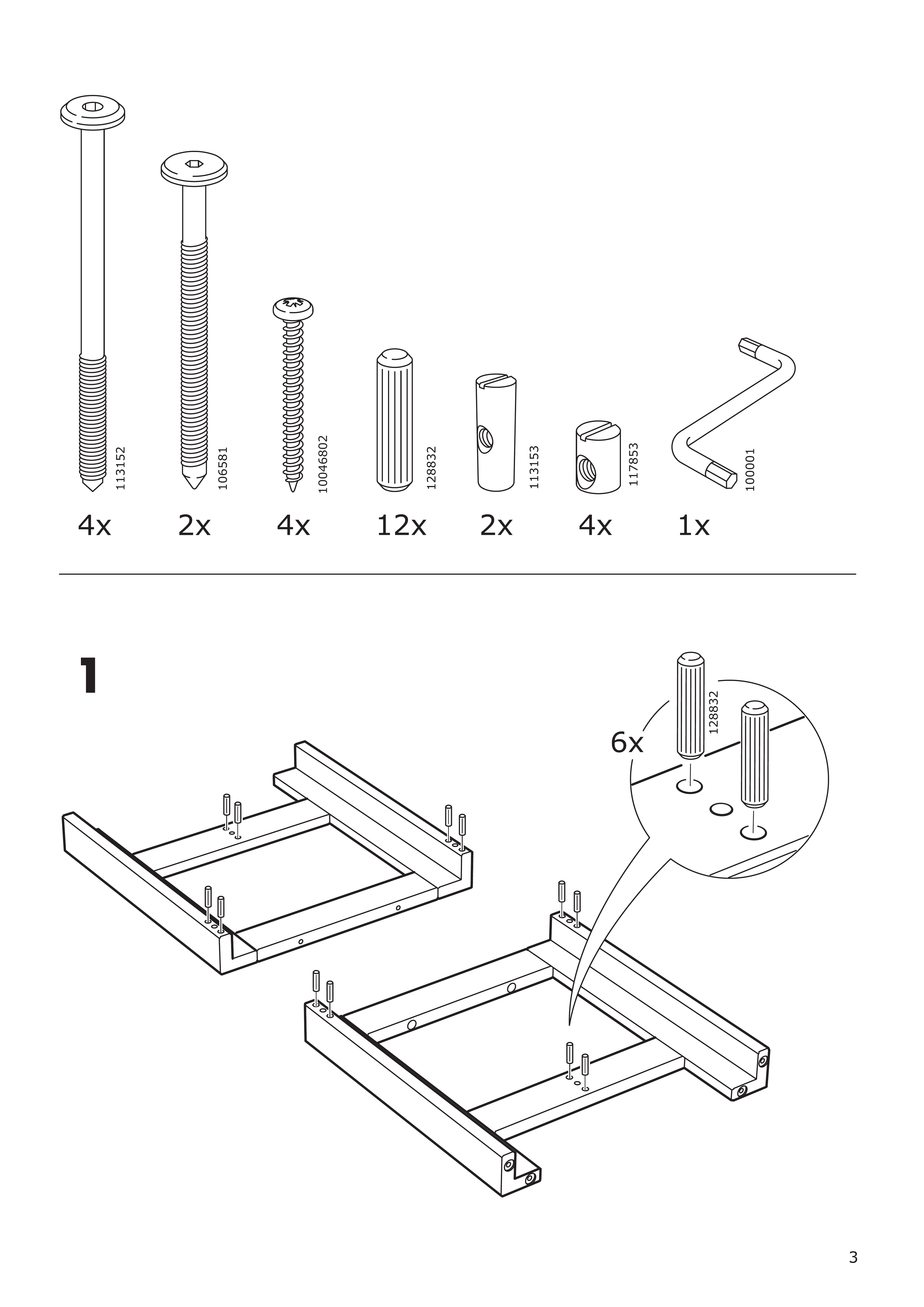}
            \includegraphics[width=0.22\textwidth]{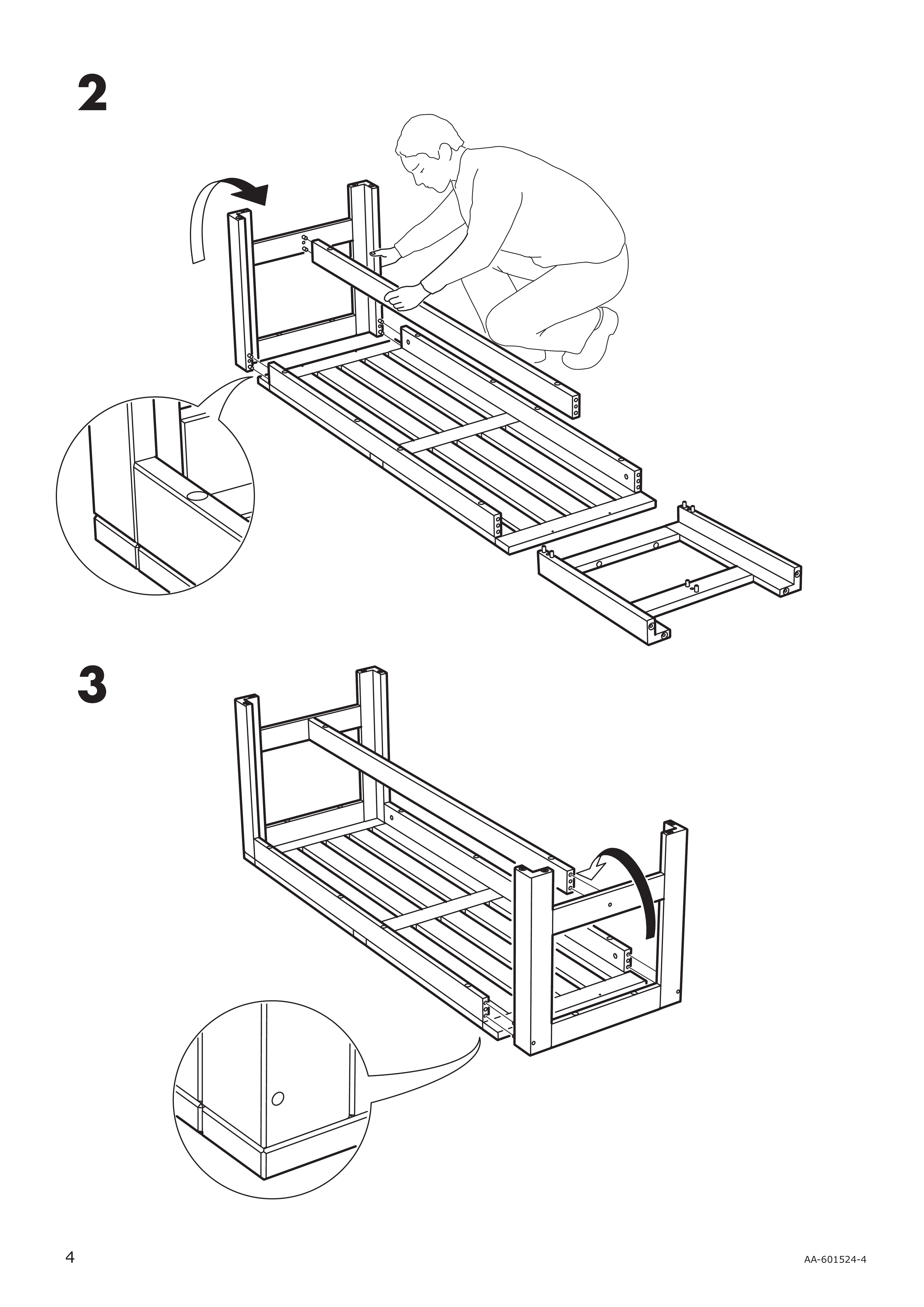}
            \includegraphics[width=0.22\textwidth]{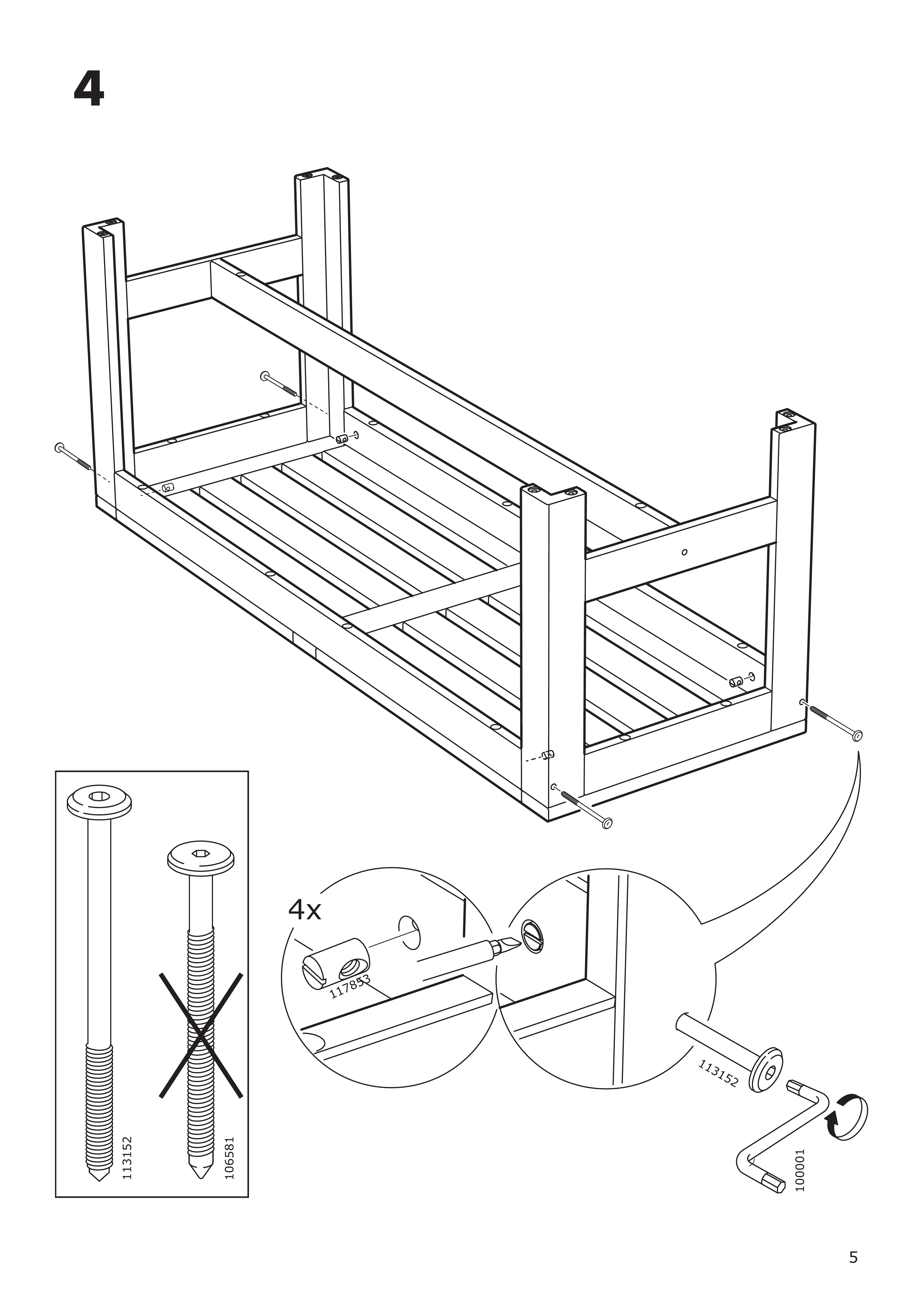}
            \includegraphics[width=0.22\textwidth]{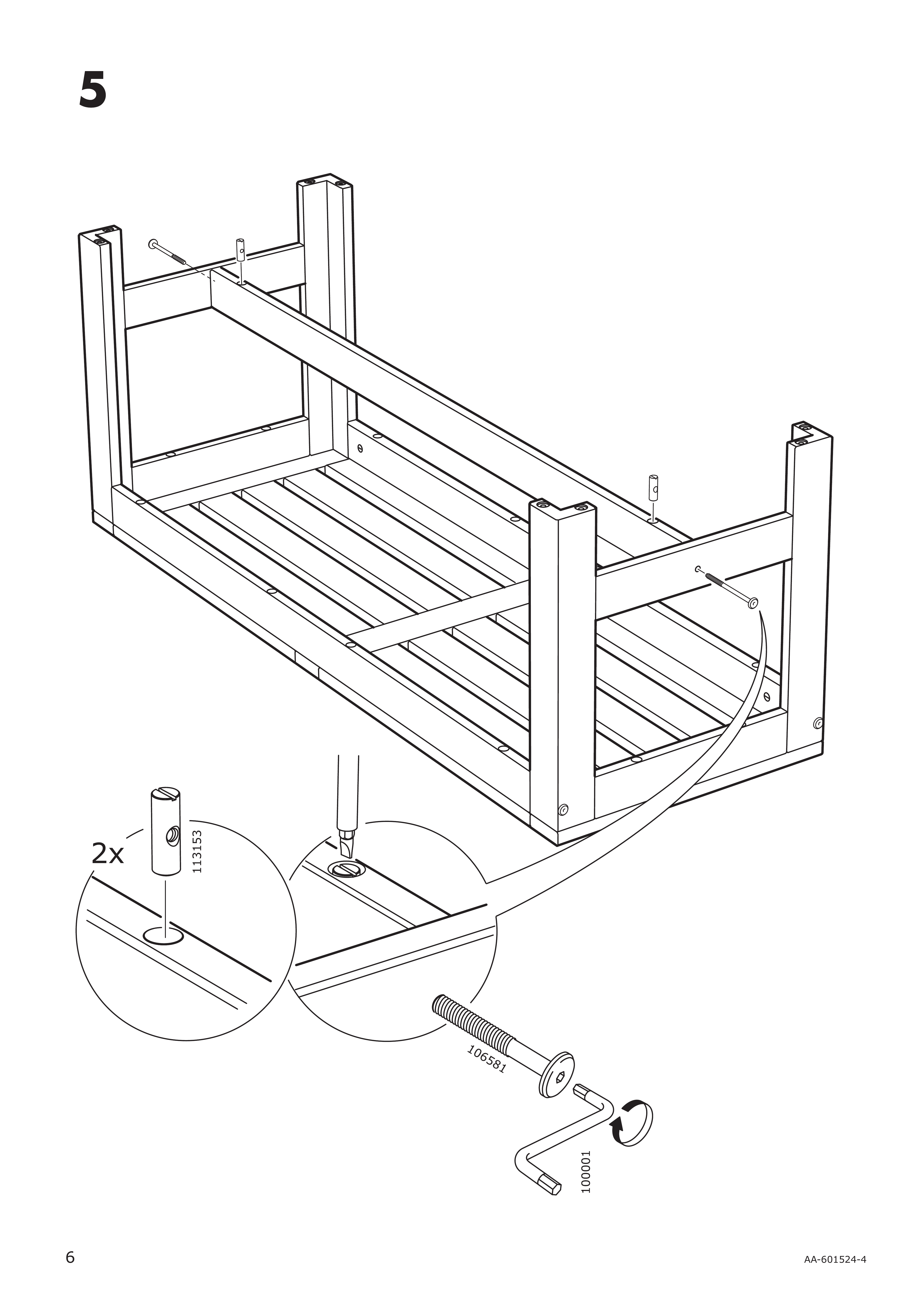}
            \includegraphics[width=0.22\textwidth]{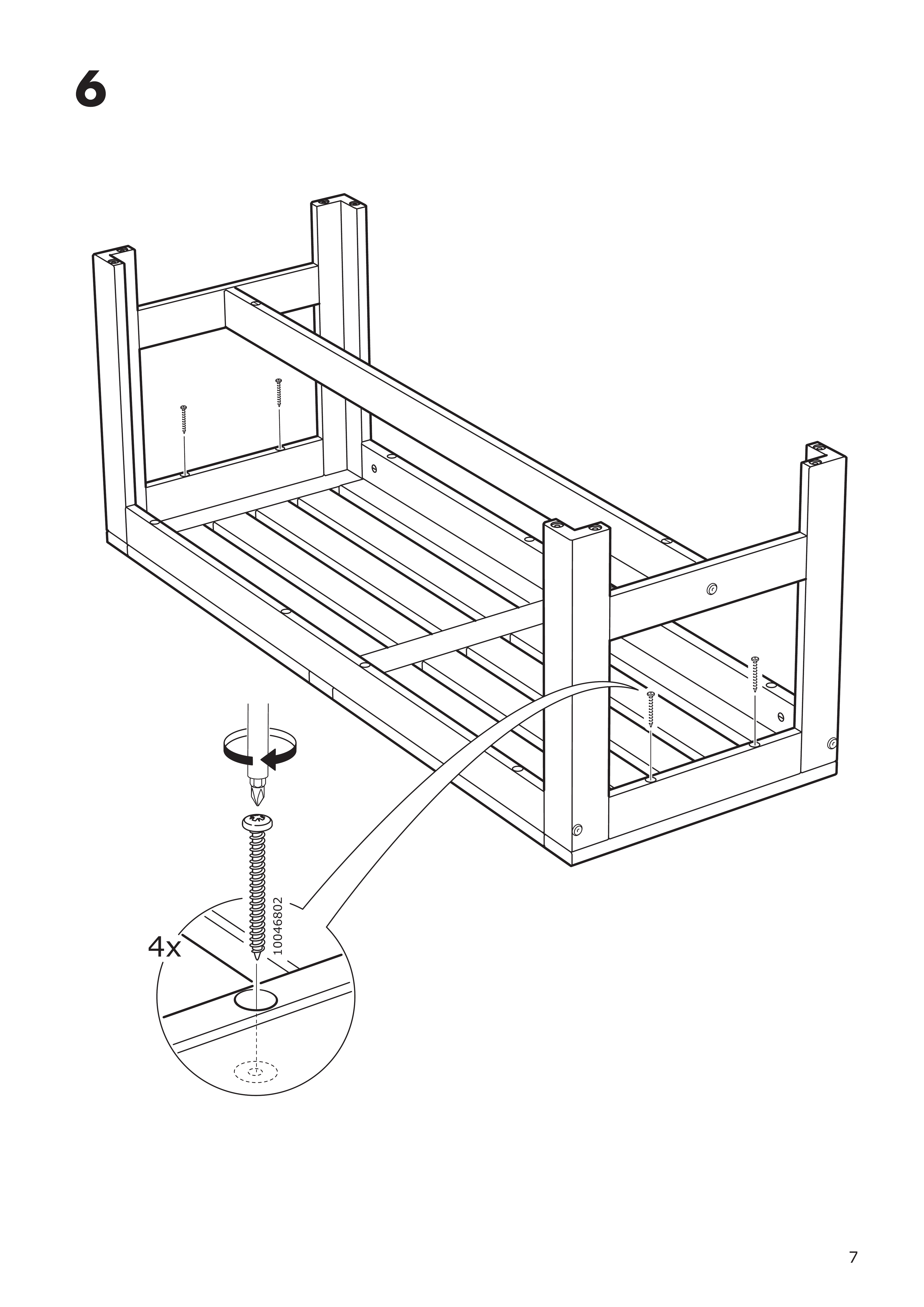}
        \end{center}\textbf{}
        \vspace{0.5em}
    \end{tcolorbox}

\begin{tcolorbox}[title=2.b) \textbf{Text Instructions} for JSON File Refinement]
You are a robot assistant responsible for assembling IKEA furniture.\newline

Your inputs include \{A\}: an rbg image of the scene consisting of furniture parts labeled with white numbers on a black background, \{B\}: a JSON file that describes the image's objects and labels, and \{C\}: a set of IKEA setup manual pages.\newline

Please note that you will only construct the piece of furniture that the manual describes.\newline

You can ignore nails and other tools in the manual and only focus on the furniture parts that exist in \{A\}: the rbg scene image.\newline

First, you are ONLY responsible for identifying the relevant materials that will be required to assemble the furniture in the image. Output a table of selected materials, with their labeled numbers and a brief explanation of why they are selected and how they are related to items on the setup manual. The table format should be JSON, and it should be really similar to \{B\}, but with an additional explanation section for each selected material and its labeled number. Hint: Usually, in 99.999\% of cases, the number of selected materials equals the number of labeled furniture parts.
\end{tcolorbox}
}]

\twocolumn[{
    \begin{tcolorbox}[title=3.a) \textbf{Image Set} for Step-By-Step Plan Generation]
        \begin{center}
            \includegraphics[width=0.3\textwidth]{Figures/scene_annotated.png}
            \includegraphics[width=0.3\textwidth]{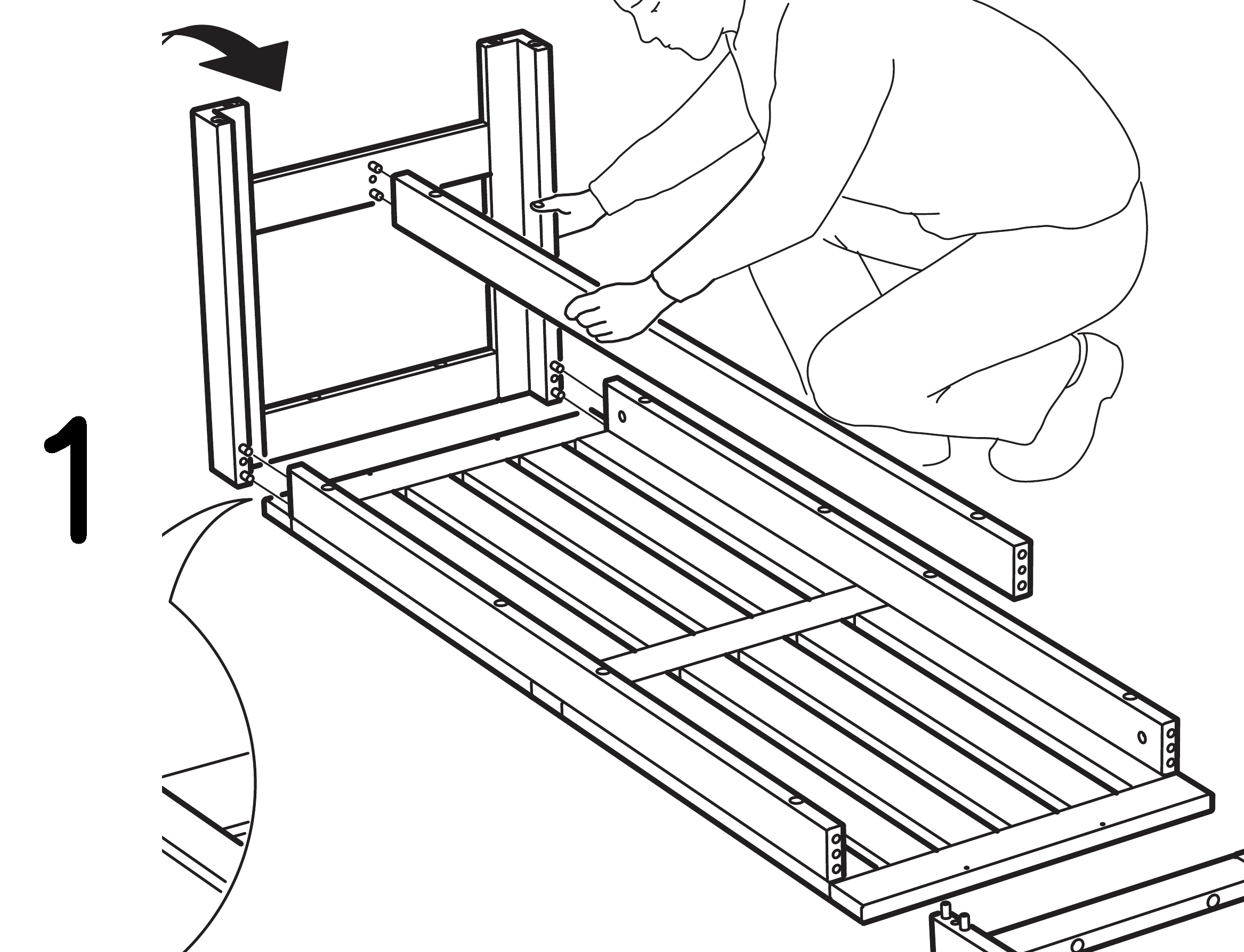}
            \includegraphics[width=0.3\textwidth]{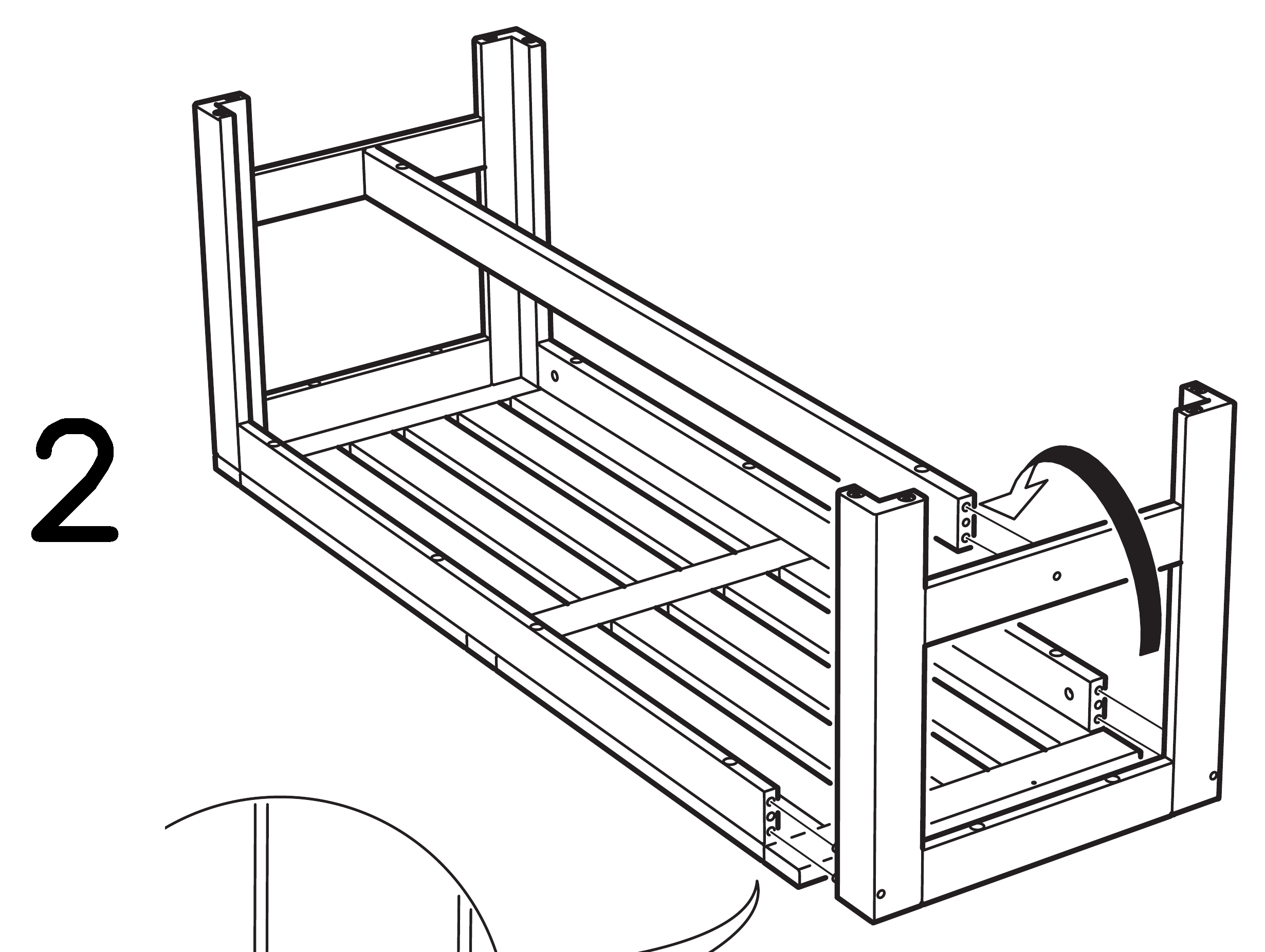}
        \end{center}\textbf{}
        \vspace{0.5em}
    \end{tcolorbox}
}]
\twocolumn[{%
\begin{tcolorbox}[title=3.b) \textbf{Text Instructions} for Step-By-Step Plan Generation]
You are a robot assistant responsible for assembling IKEA furniture. You will be responsible for creating a detailed step-by-step plan for assembling the furniture.\newline

For your input, you will receive a set of images, which represent a few pages of the setup manual containing the setup instructions for the furniture. On the left of each page, there is a rectangular section with a white background and a big, black, bolded number. This number indicates the current assembly step. On each page, we segment the furniture with different colors (the three most common are red, green, and purple, though sometimes other colors are used). The purpose of using these colors is to help you clearly identify which furniture parts are involved in each assembly step.\newline

You will also receive an rbg image of the scene consisting of furniture parts labeled with white numbers on a black background and a JSON-formatted table that describes the RGB image's objects and labels.\newline

Your new task is to carefully describe every step according to the manual. Each colored segmented furniture part should correspond to one step. Your planned steps should only describe what and how segmented furniture parts are involved; don't worry about nails and other minor tools for now. Your focus should only be on the colored segmented furniture parts. Be as specific as possible in your description.\newline

Let's think step by step: (1) count the total number of colored, segmented furniture parts. (Hint: This equals the total number of pages in the manual, with each page identified by a big, bold black number on the left.) The total number of colored, segmented furniture parts will be your total number of steps. (2) for each step, focus on one colored, segmented furniture part at a time. Describe only the furniture parts involved in that step. (3) We repeat step 2 for each remaining step until we have described all the steps. So, if there is only one page of the setup manual overlayed with mask segmentations, then there is only one step. If there are ten pages of the setup manual overlayed with mask segmentations, then there are ten steps.\newline

Here is an example of a fully constructed plan for your reference only. It has nothing to do with the current plan:\newline

\#\#\#\#\#\#\#\#\#\# assistant example start \#\#\#\#\#\#\#\#\#\#\newline
We have five input images, but one image shows furniture parts lying on a floor that we label with marks (white numbers on a black square background). Therefore, we have only four pages of the setup manual overlaid with mask segmentations. Thus, there are four total steps.\newline

\#\#\# Step 1:\newline
- **Parts Needed:** Backrest Frame (1), Seat Cushion (5)\newline
- **Instructions:**\newline
  - **Align Frame and Seat:** Connect the backrest frame (1) next to the seat cushion (5) as shown in the segmented manual.\newline

\#\#\# Step 2:\newline
- **Parts Needed:** Subassembly from Step 1, Side Leg Frame (2)\newline
- **Instructions:**\newline
  - **Position Leg Frame:** Link the first side leg frame (2) with the assembled seat and backrest combo from Step 1.\newline

\#\#\# Step 3:\newline
- **Parts Needed:** Subassembly from Step 2, Support Beam (3), Support Beam (4), Side Leg Frame (6)\newline
- **Instructions:**\newline
  - **Connect Support Beams:** Attach support beams (3), (4), and the second side leg frame (6) between the assembled frame and leg structure from Step 2. \newline

\#\#\#\#\#\#\#\#\#\# assistant example end \#\#\#\#\#\#\#\#\#\#\newline

Now it is your turn to generate a detailed step-by-step plan; here is the JSON formatted table:
\end{tcolorbox}
}]

\twocolumn[{%
\begin{tcolorbox}[title=4) Prompt for Converting Text-Formatted Plan to Tree]
You are a robot assistant responsible for assembling IKEA furnitures.\newline

Your new task is to convert a step-by-step furniture assembly instruction plan from text format into a tree format.\newline

The tree represents the stage of the furniture assembly, with lower-level nodes representing the initial and beginning stages and the upper level representing the concluding and finished stages of the furniture assembly.\newline

We treat each end node (leaf) of the tree as an atomic furniture part that we cannot further decompose. As you move up the tree, each parent node will represent two or more child nodes combined. Finally, the root node will be the completed furniture.\newline

You should clearly describe how every node is connected.\newline

We output the tree strictly as a nested list of integers without any additional comments or text.

\end{tcolorbox}
\begin{tcolorbox}[title=4) (Continued) In-Context Learning Examples for Text-Formatted Plan to Tree Prompt]
EXAMPLE INPUT 1:\newline
Here's a step-by-step assembly plan for the furniture using the provided parts:\newline

\#\#\# Step 1: Assemble Backrest and Seat\newline
- **Parts Needed:** Backrest Frame (1), Seat Cushion (5)\newline
- **Instructions:**\newline
  - Place the Backrest Frame (1) and Seat Cushion (5) adjacent as shown in their respective colors (red and green).\newline
  - Ensure the backrest is upright and securely attached to the seat.\newline

\#\#\# Step 2: Attach Side Leg Frame\newline
- **Parts Needed:** Side Leg Frame (2) and subassembly from Step 1\newline
- **Instructions:**\newline
  - Position the Side Leg Frame (2) on one side of the assembled backrest and seat structure.\newline

\#\#\# Step 3: Attach Side Leg Frame Again\newline
- **Parts Needed:** Side Leg Frame (7) and subassembly from Step 2\newline
- **Instructions:**\newline
  - Position the Side Leg Frame (7) on the other side of the assembled backrest and seat structure.\newline

\#\#\# Step 4: Connect Support Beams\newline
- **Parts Needed:** Support Beams (3, 4) and subassembly from Step 3\newline
- **Instructions:**\newline
  - Attach Support Beams (3, 4) to the inside of the Side Leg Frame, as depicted.\newline

Check the entire assembly for any loose parts and re-tighten as necessary. The chair should now be fully assembled and ready for use.\newline

EXAMPLE OUTPUT 1:\newline
'''python\newline
[
    [
        [
            [
                1,
                5
            ],
            2
        ],
        7
    ],
    3,
    4
]\newline
'''
\end{tcolorbox}

}]

\twocolumn[{%
\begin{tcolorbox}[title=4) (Continued) In-Context Learning Examples for Text-Formatted Plan to Tree Prompt]
EXAMPLE INPUT 2:\newline

\#\#\# Step 1: Connect Support Beams and Leg Frame\newline
**Parts Involved:** Support Beams (0 and 3), Leg Frame (4)\newline
- **Instructions:** Position the leg frame (4) horizontally on the floor. Align the support beams (0 and 1) vertically to connect with the leg frame. Ensure that each beam is fitted securely into the designated slots on the frame.\newline

\#\#\# Step 2: Attach Backrest Slats\newline
**Parts Involved:** Backrest Slats (2) and subassembly from Step 1\newline
- **Instructions:** Insert the backrest slats (2) into the slots on the leg frame. Ensure that the slats are facing outward and securely fitted to provide back support.\newline

\#\#\# Step 3: Connect Seat Cushion\newline
**Parts Involved:** Seat Cushion (1) and subassembly from Step 2\newline
- **Instructions:** Place the seat cushion (1) on top of the assembled frame. Align the cushion with the edges of the frame for balance and comfort.\newline

EXAMPLE OUTPUT 2:\newline
'''python\newline
[
    [
        [
            0,
            3,
            4
        ],
        2
    ],
    1
]\newline
'''\newline
EXAMPLE INPUT 3:\newline

\#\#\# Step 1: Connect Support Beams and Leg Frame\newline
**Parts Involved:** Support Beams (7, 11, 6), Leg Frame (5)\newline
- **Instructions:** Position the leg frame (5) horizontally on the floor. Align the support beams (7, 11, 6) vertically to connect with the leg frame. Ensure that each beam is fitted securely into the designated slots on the frame.\newline

\#\#\# Step 2: Attach Backrest Slats\newline
**Parts Involved:** Backrest Slats (1, 10) and subassembly from Step 1\newline
- **Instructions:** Insert the backrest slats (1, 10) into the slots on the leg frame. Ensure that the slats are facing outward and securely fitted to provide back support.\newline

\#\#\# Step 3: Connect Seat Cushion\newline
**Parts Involved:** Seat Cushion (3) and subassembly from Step 2\newline
- **Instructions:** Place the seat cushion (3) on top of the assembled frame. Align the cushion with the edges of the frame for balance and comfort.\newline

\#\#\# Step 4: Connect Support Beams and Leg Frames\newline
**Parts Involved:** Support Beams (8, 4), Leg Frames (2, 9)\newline
- **Instructions:** Position the leg frame (2, 9) horizontally on the floor. Align the support beams (8, 4) vertically to connect with the leg frame.\newline

\#\#\# Step 5: Connect Support Beams and Leg Frames\newline
**Parts Involved:** Subassembly from Step 4 and subassembly from Step 3\newline
- **Instructions:** Connect the two subassemblies together\newline

\#\#\# Step 6: Connect Support Beams and Leg Frames\newline
**Parts Involved:** Leg frame (0) and subassembly from Step 5\newline
- **Instructions:** Connect the final leg frame with the previous subassembly\newline

EXAMPLE OUTPUT 3:\newline
'''python\newline
[
    [
        [
            8,
            4,
            2,
            9
        ],
        [
            [
                [
                    7,
                    11,
                    6,
                    5
                ], 
                1,
                10 
            ],  
            3
        ]
    ]
    0]\newline
'''\\

YOUR REAL INPUT:
\end{tcolorbox}
}]

\end{document}